\documentclass[conference]{IEEEtran}
\IEEEoverridecommandlockouts    

\usepackage{multirow}
\usepackage{lipsum}
\usepackage{amsmath}
\usepackage{amssymb}
\usepackage[ruled,vlined]{algorithm2e}
\usepackage{graphicx}
\graphicspath{{./Figures/}}
\usepackage{multicol}
\usepackage{booktabs}
\usepackage{float}
\usepackage[hidelinks]{hyperref}
\usepackage{booktabs}
\usepackage{makecell}
\usepackage{graphicx}
\usepackage{float}
\usepackage{caption}
\usepackage[caption=false,font=normalsize,labelfont=sf,textfont=sf]{subfig}
\title{\LARGE \bf Competitive Multi-Operator Reinforcement Learning for Joint Pricing and Fleet Rebalancing in AMoD Systems}

\author{
 	\parbox{\textwidth}{%
 		\centering
 		Emil Kragh Toft$^{1}$, Carolin Schmidt$^{2}$, Daniele Gammelli$^{3}$, Filipe Rodrigues$^{1}$%
 	}%
	\thanks{$^{1}$Technical University of Denmark, Denmark
 		{\tt\small s233791@student.dtu.dk, rodr@dtu.dk}}%
 	\thanks{$^{2}$Technical University of Munich, Germany 
 		{\tt\small carolin.schmidt@tum.de}}%
 	\thanks{$^{3}$Stanford University, USA
 		{\tt\small gammelli@stanford.edu}}%
 }

\begin{document}
	
	\maketitle
	\thispagestyle{empty}
	\pagestyle{empty}
	
	\begin{abstract}
		Autonomous Mobility-on-Demand (AMoD) systems promise to revolutionize urban transportation by providing affordable on-demand services to meet growing travel demand. However, realistic AMoD markets will be competitive, with multiple operators competing for passengers through strategic pricing and fleet deployment. While reinforcement learning has shown promise in optimizing single-operator AMoD control, existing work fails to capture competitive market dynamics. We investigate the impact of competition on policy learning by introducing a multi-operator reinforcement learning framework where two operators simultaneously learn pricing and fleet rebalancing policies. By integrating discrete choice theory, we enable passenger allocation and demand competition to emerge endogenously from utility-maximizing decisions. Experiments using real-world data from multiple cities demonstrate that competition fundamentally alters learned behaviors, leading to lower prices and distinct fleet positioning patterns compared to monopolistic settings. Notably, we demonstrate that learning-based approaches are robust to the additional stochasticity of competition, with competitive agents successfully converging to effective policies while accounting for partially unobserved competitor strategies.
	\end{abstract}
\section{Introduction}
\label{sec:introduction}
Urban mobility systems face growing pressure from rapid urbanization, and evolving consumer expectations. The gap between private vehicle ownership and inflexible public transit has fueled the rapid growth of ride-hailing services, fundamentally transforming urban mobility \cite{Ceder02102021}. Autonomous Mobility-on-Demand (AMoD) systems promise to further revolutionize transportation by enabling affordable door-to-door service at scale \cite{WANG2024104728}. However, unlike regulated public transit monopolies, AMoD markets are likely to be competitive, with multiple operators deploying fleets and competing for passengers.

In such markets, operators must strategically set prices and position their fleet while anticipating competitor actions, e.g., a price reduction by one operator diverts passengers from others. These interactions create a complex, competitive environment where optimal policies depend on competitor behavior.

Reinforcement Learning (RL) has emerged as a powerful approach for AMoD control to derive scalable, near-optimal policies from data without requiring explicit models of demand or competitor strategies. However,  existing RL research predominantly addresses centralized, single-operator settings on vehicle rebalancing \cite{9683135, 8317908}, pricing \cite{LEI2023102848, CHEN2021103272}, or both \cite{11063454}, while realistic AMoD systems will be deployed in a competitive context.

This work investigates the impact of competition on policy learning by developing a competitive multi-operator RL framework for joint pricing and fleet rebalancing. We model two operators that simultaneously learn and adapt strategies while competing for passenger demand, which is allocated via a choice model sensitive to prices, travel times, and wages. The main contributions are:
\begin{itemize}
    \item We formulate a competitive, dual-operator AMoD control problem in which two independent operators jointly learn pricing and rebalancing policies, extending joint control RL from monopolistic to competitive markets. 
    \item We integrate a passenger choice mechanism with wage-dependent price sensitivity into the learning loop, allowing demand allocation between operators to emerge endogenously from their actions.
    \item Using real-world taxi-trip data from multiple cities, we empirically demonstrate that simultaneous learning in a competitive setting converges, and analyze how competition alters learned strategies, service quality, and market efficiency relative to monopolistic control. 
\end{itemize}

\section{Literature Review}
\label{sec:lit_review}
Early work on AMoD systems relied on queueing-theoretic models \cite{doi:10.1177/0278364915581863} and Model Predictive Control for fleet rebalancing \cite{CALAFIORE2019169}, later extended with robust optimization to handle demand uncertainty \cite{GUO2021161}. While theoretically grounded, these approaches struggle when demand deviates from assumed distributions or computational budgets are exceeded \cite{8460966}. More recently, RL has emerged as a data-driven alternative, learning policies directly through interaction without explicit demand or traffic models \cite{8317908}. Gammelli et al.\ \cite{9683135} proposed a hierarchical framework for rebalancing that combines the strengths of traditional optimization and RL, using graph neural networks to exploit the spatial structure of transportation networks. By leveraging these tools, RL scales efficiently to large-scale, real-world urban mobility scenarios \cite{tresca2025robotaxifleetcoordinationscale}. 

While these methods focus on supply, dynamic pricing complements rebalancing by shaping demand. Early pricing work adopted equilibrium-based \cite{Cachon2016TheRO} and operations-research formulations \cite{guan2021sharedmobility}, while bi-level models \cite{10.1155/2022/9120129} jointly optimize operator and passenger decisions. However, addressing pricing or rebalancing in isolation fails to exploit their synergies \cite{11063454}. In particular, a recent RL-based framework for joint rebalancing and pricing has shown that joint control consistently outperforms separate strategies \cite{11063454}. 

Multi-operator competition has been studied through game-theoretic pricing models \cite{yang2025intraday}, agent-based simulation \cite{10.3389/ffutr.2022.915219}, and quality-of-service competition frameworks \cite{10.1109/ITSC57777.2023.10421926}. However, these works rely on analytical equilibrium or static models and address either pricing or fleet management in isolation. While reinforcement learning has been explored in single-operator AMoD contexts, this work addresses the competitive multi-operator setting 
to jointly learn competitive rebalancing and pricing policies with endogenous, price-responsive demand. 
Specifically, by formulating a dual-operator RL framework where competing operators simultaneously learn policies while competing for demand, this work fills the gap and studies the impact of competition on RL-based policy learning. 

\section{Design and Implementation}
We represent the AMoD environment as a directed graph $\mathcal{G} = (\mathcal{V}, \mathcal{E})$, where $\mathcal{V}$ comprises $N_v$ spatial regions, each centered on a designated pickup/drop-off station. The total fleet is partitioned between two independent operators controlling $M_0$ and $M_1$ autonomous vehicles, respectively. Although we focus on the dual-operator case for clarity, the framework extends trivially to more operators. Both operators act within a synchronized discrete time horizon $\mathcal{T} = \{1,\dots,T\}$ with time steps of 3 minutes. Movement from region $i$ to $j$ follows the shortest path, requiring $\tau_{ij}\in\mathbb{Z}_+$ time steps and incurring a time-variant operational cost $c_{ij}^t$.

Available supply at station $i$ at time $t$ is characterized by the idle-vehicle counts $m_{i,0}^t$ and $m_{i,1}^t$ of each fleet. On the demand side, passengers arriving at origin~$i$ evaluate both AMoD operators and an alternative transportation mode via a discrete choice model: a utility $U$ is computed for each option as a function of trip price, passenger salary, and estimated travel time, mapped to choice probabilities, and sampled from a categorical distribution. Upon selecting an operator, passengers enter a first-come, first-served (FCFS) queue. A maximum waiting threshold of 6 minutes is imposed; if no vehicle is dispatched within this window, the passenger exits the system.

Building on this structure, the following sections formalize the operator interaction. We first cast the pricing and rebalancing task as a Markov Decision Process (MDP), then detail a a graph neural network architecture designed to capture the spatial dependencies inherent in dual-operator learning.

\subsection{Competitive AMoD Control}
\label{sec:AMOD Control Min}

We study a dual-operator AMoD setting in which two independent operators each seek to maximize their own profit within a shared environment. The control architecture follows the hierarchical formulation of~\cite{9683135, 11063454}, enabling scalability to real-world urban scenarios \cite{tresca2025robotaxifleetcoordinationscale}. We adapt this framework to a competitive setting, as a means to study the impact of competition on RL-based policy learning. 
In practice, the control architecture introduces three steps, illustrated in the appendix.
Specifically, at each time step $t$, the simulation proceeds in three steps: \textbf{(1) Pricing \& rebalancing:} each operator's policy produces origin-based price scalars and a desired idle-vehicle distribution, where price scalars are then converted to fares by multiplication with OD-base prices estimated from historical data. \textbf{(2) Demand assignment:} the stochastic choice model generates passenger requests and assigns them to operators via the price-dependent choice model. Assigned passengers queue at their departure region and await service in FCFS order, subject to a system-wide maximum waiting time. \textbf{(3) Rebalancing execution:} each operator's desired distribution is mapped to rebalancing actions by solving a minimal-cost flow problem (see appendix). Flows are executed independently per operator, vehicle states and queues are updated, and the system transitions to $t{+}1$, returning the new state and rewards.

\subsection{The AMoD Control Problem as an MDP}
\label{sec: AMOD_MDP}

We formulate multi-operator AMoD control as an MDP with components $(\mathcal{S}, \mathcal{A}, \mathcal{P}, \mathcal{R})$ defined below.

\textit{State space} ($\mathcal{S}$): The state $\mathbf{s}_{t,o}$ of operator $o$ at time $t\in\mathcal{T}$ comprises: the network adjacency matrix $\mathbf{A}$, own idle vehicles per region $i$ as $m_{i,o}^t\in[0,M_o]$, vehicles en route over a planning horizon $H_p$, $\{m_{i,o}^{t'}\}_{t'=\tau,\dots,\tau+H_p}$, own and competitor prices from last time step $p_{i,j,o}^{t-1}$, $p_{i,j,o'}^{t-1}$ for all $i,j\in\mathcal{V}$, $o'\neq o$, and own queue lengths $q_{i,o}^t$ and own demand for last time step $d_{i,o}^{t-1}$ for all $i\in\mathcal{V}$. Operators do not share demand or vehicle-location data, but can observe the competitor's prices.

\textit{Action space} ($\mathcal{A}$): Each operator~$o$ outputs, origin-based price scalars $\rho_{i,o}^t\in(0,1]$ and a desired idle-vehicle distribution $w_{i,o}^t\in[0,1]$ with $\sum_{i=1}^{N_v}w_{i,o}^t=1$, with the latter being used to solve a minimum-cost rebalancing problem. Origin-based price scalars are chosen based on prior work \cite{11063454}, which shows faster convergence and comparable performance to OD-based price scalars. The joint action is $\mathbf{a}_{t,o}=[\mathbf{w}_{t,o},\,\mathbf{\rho}_{t,o}]$ with $\mathbf{w}_{t,o}=[w_{1,o}^t,\dots,w_{N_v,o}^t]$ and $\mathbf{\rho}_{t,o}=[\rho_{1,o}^t,\dots,\rho_{N_v,o}^t]$. The OD fare then calculated as follows:
\begin{equation}\label{price_eq}
    p_{i,j,o}^t = \beta \cdot \rho_{i,o}^t \cdot \overline{p}_{i,j}^t,
\end{equation}
where $\overline{p}_{i,j}^t$ is a historical reference price (derived from data) and $\beta$ upper-bounds the price level (we set $\beta=2$ in our experiments, giving $p_{i,j,o}^t\in(0,\,2\overline{p}_{i,j}^t]$).

\textit{Reward} ($\mathcal{R}$): Each operator maximizes its own profit, defined as trip revenue minus operational costs:
\begin{equation}
    r_{o} = \sum_{i,j\in\mathcal{V}} x_{i,j,o}^t\bigl(p_{i,j,o}^t - c_{i,j,o}^t\bigr) - \sum_{(i,j)\in\mathcal{E}} y_{i,j,o}^t\, c_{i,j,o}^t,
\end{equation}
where $x_{i,j,o}^t$ denotes passengers served, $y_{i,j,o}^t$ vehicles rebalanced, and $c_{i,j,o}^t$ the per-vehicle travel cost from $i$ to $j$.

\textit{Dynamics} ($\mathcal{P}$): State transitions are driven by stochastic demand, queue updates, and vehicle-availability conservation. The adjacency matrix $\mathbf{A}$ is exogenous and static.

\subsubsection*{Demand Generation and Mode Choice}
The choices of each passenger~$k$ are sampled between operator~0, operator~1, or an alternative option ($\emptyset$) via a Multinomial Logit model with utility
\begin{equation}
    U^t_{k,i,j,o} = \beta_0 - \beta_t \cdot v_k \cdot \tau^t_{i,j} - \frac{\bar{v}}{v_k}\, p^t_{i,j,o},
    \label{eq:utility}
\end{equation}
where $\beta_0$ is a baseline preference, $\beta_t$ the marginal disutility of travel time (set to $0.71$, reflecting a 29\% undervaluation of time relative to monetary cost, consistent with~\cite{khulbe2023probabilistic}), $v_k$ the passenger's hourly wage, and $\bar{v}$ the scenario-wide average wage. The ratio $\bar{v}/v_k$ captures income effects. The alternative-option utility is normalized to zero. For each OD-pair $(i,j)$ and time $t$, a reference demand $\bar{d}_{i,j}^t$ and price $\bar{p}_{i,j}^t$ are sampled from historical data \cite{gammelli2022graphmetareinforcementlearningtransferable, illinoisdatabankIDB-9610843}. The potential demand pool is scaled to $2\bar{d}_{i,j}^t$, and $\beta_0$ calibrated such that the total operator demand at historic fare prices is equal the reference demand. The choice model sensitivity to price scalars is shown in the appendix. Wage data are derived from US Census Bureau income statistics~\cite{uscensus_s1901_2013}, with wages for the San Francisco scenario being adjusted using the national US inflation rate from 2008 to 2011 to obtain 2008-equivalent estimates from 2011 data. 

\subsubsection*{Queue and Vehicle Dynamics}
The queue for operator~$o$ in region~$i$ evolves as
\begin{equation}
    q_{i,o}^{t} = q_{i,o}^{t-1} + \sum_{j\in\mathcal{V}\setminus\{i\}} \bigl(d_{i,j,o}^{t} - x^t_{i,j,o}\bigr),
\end{equation}
balancing new arrivals against successful matches. Since autonomous vehicles comply fully with dispatch directives, the idle-vehicle count updates as
\begin{equation}
    m_{i,o}^t = m_{i,o}^{t-1} + \sum_{j\in\mathcal{V}\setminus\{i\}} v_{j,i,o}^{\text{arr},t} - \sum_{j\in\mathcal{V}\setminus\{i\}} \bigl(x_{i,j,o}^{t-1} + y_{i,j,o}^{t-1}\bigr),
\end{equation}
where $v_{j,i,o}^{\text{arr},t}$ counts vehicles arriving at $i$ from $j$ at time $t$ (completing passenger trips or rebalancing manoeuvres from prior intervals), and the outflow comprises matched departures $x_{i,j,o}^{t-1}$ and rebalancing dispatches $y_{i,j,o}^{t-1}$.

\subsection{Model Architecture}
\label{sec:Arch}
At a high level, the objective for each operator is to learn a policy $\pi(a_t|s_t)$ that maximizes their expected cumulative reward through interactions with the MDP. In our setup, each operator~$o$ maintains an independent actor $\pi_{\theta_o}(\cdot\mid\mathbf{s}_t)$ and critic $V_{\phi_o}(\mathbf{s}_t)$ with no parameter sharing between operators. The networks take the state $\mathbf{s}_t$ (Section~\ref{sec: AMOD_MDP}) as input and employ a a convolutional neural network (GCN) encodings~\cite{9683135, 11063454} followed by fully connected layers (see appendix). The critic aggregates node embeddings via global summation to produce a scalar value $V_{\phi_o}(\mathbf{s}_t)$. The actor outputs strictly positive parameters (via Softplus) that define the stochastic policy. For joint pricing and rebalancing, each region~$i$ yields parameters $\alpha_i^t$ and $\beta_i^t$ for a Beta distribution from which the origin-based price scalar is sampled: $p^t_{i,o} \sim \text{Beta}(\alpha_i, \beta_i)$, and a concentration scalar $\gamma_i^t$ that collectively form $\boldsymbol{\gamma}^t\in\mathbb{R}^{N_v}$, parameterizing a Dirichlet distribution from which the desired idle-vehicle shares are sampled: $\mathbf{w}^t_o \sim \text{Dirichlet}(\boldsymbol{\gamma}^t)$. For pricing-only or rebalancing-only control, the actor outputs only the Beta or Dirichlet parameters, respectively.

\section{Experiments}
\label{sec:Experiments}
This section empirically evaluates the proposed framework along two dimensions: (i) performance across San Francisco, Washington DC, and NYC Manhattan South, contrasting monopolistic single-operator control against competitive dual-operator scenarios under rebalancing-only, pricing-only, and joint control modes; and (ii) sensitivity analyses on fleet size, asymmetric fleet distributions, and regional wage heterogeneity and their effects on equilibrium outcomes.

All experiments use a discrete-time simulation environment adapted from \cite{gammelli2022graphmetareinforcementlearningtransferable}, spanning the peak evening period (19:00--20:00).
  Demand is modeled via a Poisson distribution where the rate parameter is derived from real-world taxi trip data \cite{gammelli2022graphmetareinforcementlearningtransferable, illinoisdatabankIDB-9610843}. Operators are trained using A2C with the GCN-based architecture described in Section~\ref{sec:Arch}, trained to convergence and evaluated over 10 test runs. We report averages and standard deviations in parentheses. 
  \footnote{Supplementary material, code and hyperparameter at \url{https://github.com/emto17ab/dual_agent_rl}}

\subsection{Multi-City Performance Analysis}
\label{sec:exp_multicity}
We evaluate performance across three urban environments with distinct characteristics, summarized in Table~\ref{tab:scenarios}. Spatial demand variability is measured by the coefficient of variation (CV), defined as the ratio of the standard deviation to the mean of regional demand. We further report two non-learned benchmark gaps: The no control (``NC'') baseline, corresponding to system performance without rebalancing or pricing interventions, and a uniform distribution heuristic (``UD''), which aims to distribute idle vehicles evenly across the network under fixed prices. 
The reward gap and served gap measure the relative improvement of equal-distribution over no-control in reward and served passengers, serving as an approximation for the improvement gained by rebalancing. 

\begin{table}[!h]
\centering
\caption{Scenario characteristics for comparative analysis.}
\label{tab:scenarios}
\resizebox{\columnwidth}{!}{%
\begin{tabular}{l c c c c c c c}
\toprule
City & \makecell{Hourly\\Wage} & Nodes & Veh. & Demand & \makecell{Reward\\Gap} & \makecell{Served\\Gap} & CV \\
\midrule
\makecell[l]{San\\Francisco} & \$17.76 & 10 & 374 & 5490 & 46.72\% & 63.04\% & 1.31 \\
\makecell[l]{Washington\\DC} & \$25.26 & 18 & 1096 & 16881 & 15.14\% & 48.13\% & 1.26 \\
\makecell[l]{NYC Man.\\South} & \$22.77 & 12 & 650 & 21270 & 10.03\% & 20.83\% & 0.69 \\
\bottomrule
\multicolumn{8}{l}{\footnotesize \textit{Note:} Veh.: No. Vehicles; CV: Coefficient of Variation of Demand.} \\
\end{tabular}%
}
\end{table}

\subsubsection{Single-Operator Monopolistic Performance}
Tables~\ref{tab:policy_performance} and \ref{tab:performance_metrics} present total rewards for different control strategies in the single-operator setting. Beyond the no-control and uniform distribution heursitic, learning-based methods include rebalancing-only, pricing-only and a joint policy that controls both decisions simultaneously. The joint policy outperforms all baselines and single-mode policies across all three cities. In San Francisco, joint control achieves a 22.4\% improvement over rebalancing alone and a 42.4\% improvement over pricing alone. The magnitude of this improvement decreases in cities with lower demand variability: the gain over rebalancing drops to 5.3\% in Washington DC, and becomes negligible (0.02\%) in NYC Manhattan South. Dynamic pricing, under both pricing-only and joint control policies, leads to a monopolistic increase in prices in both Washington DC and NYC Man South, decreasing the number of served passengers compared to rebalancing with fixed prices. 
\begin{table}
\centering
\caption{Single-operator (monopolistic) profits [\$].  Best policy in bold. We report averages across 10 test runs and std. dev in parentheses}
\label{tab:policy_performance}
\resizebox{\columnwidth}{!}{%
\begin{tabular}{l c c c c c}
\toprule
City & NC & UD & Reb. & Pricing & Joint \\
\midrule
San Francisco & \makecell{6345.49 \\ \footnotesize (251.60)} & \makecell{9470.06 \\ \footnotesize (336.71)} & \makecell{10037.29 \\ \footnotesize (337.03)} & \makecell{8629.30 \\ \footnotesize (193.96)} & \makecell{\textbf{12286.03} \\ \footnotesize \textbf{(166.74)}} \\[0.5em]
Washington DC & \makecell{13153.97 \\ \footnotesize (324.58)} & \makecell{15630.03 \\ \footnotesize (388.50)} & \makecell{15976.21 \\ \footnotesize (395.37)} & \makecell{13972.47 \\ \footnotesize (301.58)} & \makecell{\textbf{16816.04} \\ \footnotesize \textbf{(287.07)}} \\[0.5em]
NYC Man. South & \makecell{16283.02 \\ \footnotesize (487.67)} & \makecell{18221.62 \\ \footnotesize (260.07)} & \makecell{18499.21 \\ \footnotesize (291.51)} & \makecell{17004.39 \\ \footnotesize (504.23)} & \makecell{\textbf{18503.52} \\ \footnotesize \textbf{(324.97)}} \\[0.5em]
\bottomrule
\end{tabular}%
}
\end{table}

\begin{table*}[t]
\centering
\caption{Single-operator (monopolistic) performance..``Wait time'' corresponds to passenger wait
time. Best results are in bold.}
\label{tab:performance_metrics}
\resizebox{0.9\textwidth}{!}{%
\begin{tabular}{l ccc ccc ccc}
\toprule
& \multicolumn{3}{c}{San Francisco} & \multicolumn{3}{c}{Washington DC} & \multicolumn{3}{c}{NYC Man. South} \\
\cmidrule(lr){2-4} \cmidrule(lr){5-7} \cmidrule(lr){8-10}
Policy & Reb. & Pricing & Joint & Reb. & Pricing & Joint & Reb. & Pricing & Joint \\
\midrule
Reward [\$] & \makecell{10037.29 \\ \footnotesize (337.03)} & \makecell{8629.30 \\ \footnotesize (193.96)} & \makecell{\textbf{12286.03} \\ \footnotesize \textbf{(166.74)}} & \makecell{15976.21 \\ \footnotesize (395.37)} & \makecell{13972.47 \\ \footnotesize (301.58)} & \makecell{\textbf{16816.04} \\ \footnotesize \textbf{(287.07)}} & \makecell{18499.21 \\ \footnotesize (291.51)} & \makecell{17004.39 \\ \footnotesize (504.23)} & \makecell{\textbf{18503.52} \\ \footnotesize \textbf{(324.97)}} \\[0.5em]
\cmidrule(lr){1-10}
Rebalancing Costs  [\$] & \makecell{877.08 \\ \footnotesize (25.30)} & \makecell{---} & \makecell{650.04 \\ \footnotesize (17.36)} & \makecell{3475.65 \\ \footnotesize (152.04)} & \makecell{---} & \makecell{2975.85 \\ \footnotesize (83.57)} & \makecell{1436.70 \\ \footnotesize (85.48)} & \makecell{---} & \makecell{1712.25 \\ \footnotesize (101.98)} \\[0.5em]
\cmidrule(lr){1-10}
Rebalance Trips & \makecell{523.70 \\ \footnotesize (12.59)} & \makecell{---} & \makecell{384.00 \\ \footnotesize (13.70)} & \makecell{1085.80 \\ \footnotesize (34.25)} & \makecell{---} & \makecell{955.50 \\ \footnotesize (25.45)} & \makecell{295.90 \\ \footnotesize (20.40)} & \makecell{---} & \makecell{361.10 \\ \footnotesize (23.56)} \\[0.5em]
\cmidrule(lr){1-10}
Price Scalar & \makecell{---} & \makecell{0.75 \\ \footnotesize (0.00)} & \makecell{0.88 \\ \footnotesize (0.00)} & \makecell{---} & \makecell{1.20 \\ \footnotesize (0.00)} & \makecell{1.02 \\ \footnotesize (0.00)} & \makecell{---} & \makecell{1.07 \\ \footnotesize (0.00)} & \makecell{1.02 \\ \footnotesize (0.00)} \\[0.5em]
\cmidrule(lr){1-10}
Wait time [min.] & \makecell{0.85 \\ \footnotesize (0.14)} & \makecell{1.49 \\ \footnotesize (0.08)} & \makecell{1.57 \\ \footnotesize (0.15)} & \makecell{1.03 \\ \footnotesize (0.09)} & \makecell{1.25 \\ \footnotesize (0.15)} & \makecell{0.64 \\ \footnotesize (0.09)} & \makecell{1.91 \\ \footnotesize (0.06)} & \makecell{1.48 \\ \footnotesize (0.08)} & \makecell{1.71 \\ \footnotesize (0.10)} \\[0.5em]
\cmidrule(lr){1-10}
Queue & \makecell{2.03 \\ \footnotesize (0.34)} & \makecell{4.44 \\ \footnotesize (0.26)} & \makecell{5.40 \\ \footnotesize (0.41)} & \makecell{5.56 \\ \footnotesize (0.44)} & \makecell{4.51 \\ \footnotesize (0.42)} & \makecell{3.03 \\ \footnotesize (0.39)} & \makecell{14.20 \\ \footnotesize (0.48)} & \makecell{9.09 \\ \footnotesize (0.40)} & \makecell{12.07 \\ \footnotesize (0.65)} \\[0.5em]
\cmidrule(lr){1-10}
Served Demand & \makecell{960.90 \\ \footnotesize (30.48)} & \makecell{816.60 \\ \footnotesize (15.14)} & \makecell{1273.70 \\ \footnotesize (16.36)} & \makecell{4182.40 \\ \footnotesize (59.60)} & \makecell{2309.50 \\ \footnotesize (49.93)} & \makecell{3860.30 \\ \footnotesize (44.29)} & \makecell{3571.10 \\ \footnotesize (36.75)} & \makecell{2777.80 \\ \footnotesize (78.12)} & \makecell{3528.10 \\ \footnotesize (39.21)} \\[0.5em]
\cmidrule(lr){1-10}
Total Demand & \makecell{1092.10 \\ \footnotesize (42.86)} & \makecell{1298.90 \\ \footnotesize (37.03)} & \makecell{1685.60 \\ \footnotesize (35.72)} & \makecell{4753.00 \\ \footnotesize (76.79)} & \makecell{2975.40 \\ \footnotesize (43.52)} & \makecell{4129.40 \\ \footnotesize (54.38)} & \makecell{4705.70 \\ \footnotesize (73.36)} & \makecell{3590.10 \\ \footnotesize (73.90)} & \makecell{4420.00 \\ \footnotesize (77.77)} \\[0.5em]
\bottomrule
\end{tabular}%
}
\end{table*}

\subsubsection{Dual-Operator Competitive Performance}
Table~\ref{tab:dual_agent_results} presents total rewards for the dual-operator setting, where, unlike the single-operator case, no single control mode dominates across all cities. In San Francisco, joint control achieves the highest reward (10,415.\$), while Washington DC favors rebalancing (15,620\$). Notably, the rebalancing-only policy serves as a fixed-price benchmark and is unrealistic in practice, as both operators share identical prices and are not in price competition. In NYC Manhattan South, pricing-only control achieves the highest reward (18,632\$), surpassing both alternative modes. This variation suggests that in high-variability environments, fleet positioning is the primary competitive lever, whereas in stable, high-density settings, pricing can effectively function as a mechanism for implicit rebalancing. Furthermore, the inferior performance of the joint policy in NYC South highlights the higher complexity of the joint control problem in the competitive setting, where rebalancing as a long-horizon decision, needs to account for the uncertainty of future demand, that is heavily influenced by the unobserved strategy of the competitor. 

\begin{table}[H]
\centering
\caption{Dual-operator combined profits [\$].  Best policy in bold.}
\label{tab:dual_agent_results}
\resizebox{0.45\textwidth}{!}{%
\begin{tabular}{l c c c c c}
\toprule
City & NC & UD & Reb. & Pricing & Joint \\
\midrule
San Francisco & \makecell{6391.2 \\ \footnotesize (232.4)} & \makecell{9352.6 \\ \footnotesize (258.9)} & \makecell{10065.9 \\ \footnotesize (248.7)} & \makecell{8919.0 \\ \footnotesize (160.4)} & \makecell{\textbf{10415.1} \\ \footnotesize \textbf{(178.8)}} \\[0.5em]
Washington DC & \makecell{13172.3 \\ \footnotesize (344.6)} & \makecell{15113.5 \\ \footnotesize (325.8)} & \makecell{\textbf{15619.5} \\ \footnotesize \textbf{(325.7)}} & \makecell{12820.8 \\ \footnotesize (318.2)} & \makecell{13460.6 \\ \footnotesize (247.6)} \\[0.5em]
NYC Man. South & \makecell{16048.4 \\ \footnotesize (416.6)} & \makecell{17652.0 \\ \footnotesize (277.5)} & \makecell{18133.7 \\ \footnotesize (330.8)} & \makecell{\textbf{18631.6} \\ \footnotesize \textbf{(376.7)}} & \makecell{17481.1 \\ \footnotesize (258.7)} \\[0.5em]
\bottomrule
\end{tabular}%
}
\end{table}
Table~\ref{tab:performance_metrics_dual} provides detailed metrics for the dual-operator setting. Several patterns are worth highlighting. First, competition drives prices downward relative to the monopolistic case. Second, waiting times generally increase in the dual-operator setting, reflecting the inefficiency of fragmented fleet management. Third, profit splits between operators are approximately balanced, with typical differences under 2\%.

\begin{table*}[!t]
\centering
\caption{Dual-operator performance metrics. Best-performing policies are in bold.}
\label{tab:performance_metrics_dual}
\resizebox{\textwidth}{!}{%

\begin{tabular}{l ccc ccc ccc}
\toprule
& \multicolumn{3}{c}{San Francisco} & \multicolumn{3}{c}{Washington DC} & \multicolumn{3}{c}{NYC Man. South} \\
\cmidrule(lr){2-4} \cmidrule(lr){5-7} \cmidrule(lr){8-10}
Policy & Reb. & Pricing & Joint & Reb. & Pricing & Joint & Reb. & Pricing & Joint \\
\midrule
Total Reward  [\$] & 10065.9 (248.7) & 8919.0 (160.4) & \textbf{10415.1 (178.8)} & \textbf{15619.5 (325.7)} & 12820.8 (318.2) & 13460.6 (247.6) & 18133.7 (330.8) & \textbf{18631.6 (376.7)} & 17481.1 (258.7) \\
Reward Operator 0 & 5071.5 (178.5) & 4191.5 (77.1) & 5190.7 (116.7) & 7795.7 (233.7) & 6386.6 (189.6) & 6726.0 (135.8) & 9010.8 (242.6) & 9351.6 (161.4) & 8717.0 (249.1) \\
Reward Operator 1 & 4994.4 (159.8) & 4727.6 (149.5) & 5224.4 (126.3) & 7823.8 (117.2) & 6434.2 (255.9) & 6734.6 (171.4) & 9123.0 (341.4) & 9280.1 (265.1) & 8764.1 (223.5) \\
\cmidrule(lr){1-10}
Total Reb. Costs  [\$]& 1006.8 (24.0) & --- & 549.1 (20.7) & 4120.8 (120.2) & --- & 3368.1 (125.2) & 1600.2 (95.4) & --- & 1379.0 (83.7) \\
Reb. Costs Operator 0 & 498.5 (19.3) & --- & 296.7 (10.6) & 2094.8 (88.5) & --- & 1682.7 (72.1) & 813.5 (73.6) & --- & 699.6 (76.7) \\
Reb. Costs Operator 1 & 508.3 (19.0) & --- & 252.4 (13.3) & 2026.0 (59.2) & --- & 1685.4 (83.3) & 786.8 (107.0) & --- & 679.4 (63.1) \\
\cmidrule(lr){1-10}
Total Reb. Trips & 593.8 (14.4) & --- & 323.8 (13.5) & 1221.9 (47.4) & --- & 911.1 (37.2) & 338.4 (21.2) & --- & 284.7 (19.0) \\
Reb. Trips Operator 0 & 294.0 (11.0) & --- & 172.3 (5.3) & 618.2 (37.0) & --- & 457.2 (20.4) & 173.5 (18.2) & --- & 144.9 (17.8) \\
Reb. Trips Operator 1 & 299.8 (9.7) & --- & 151.5 (9.8) & 603.7 (15.3) & --- & 453.9 (21.6) & 164.9 (24.2) & --- & 139.8 (14.3) \\
\cmidrule(lr){1-10}
Total Served Demand & 972.3 (23.4) & 865.8 (14.4) & 1378.9 (17.1) & 4239.5 (46.8) & 2892.8 (67.6) & 4414.4 (33.0) & 3535.8 (39.4) & 3337.6 (52.3) & 3561.1 (29.0) \\
Served Demand  & 488.4 (16.1) & 403.6 (6.7) & 694.5 (12.3) & 2124.2 (34.5) & 1447.5 (43.7) & 2205.9 (18.5) & 1758.1 (29.4) & 1674.3 (19.1) & 1773.9 (26.1) \\
Served Demand  & 483.9 (15.1) & 462.2 (13.8) & 684.4 (11.7) & 2115.3 (16.7) & 1445.3 (56.6) & 2208.5 (21.4) & 1777.7 (39.5) & 1663.3 (38.7) & 1787.2 (24.0) \\
\cmidrule(lr){1-10}
Price Scalar Operator 0 & --- & 0.72 (0.00) & 0.64 (0.00) & --- & 0.97 (0.00) & 0.90 (0.00) & --- & 0.95 (0.00) & 0.96 (0.00) \\
Price Scalar Operator 1 & --- & 0.72 (0.00) & 0.67 (0.00) & --- & 0.98 (0.00) & 0.90 (0.00) & --- & 0.95 (0.01) & 0.96 (0.00) \\
\cmidrule(lr){1-10}
Wait time Operator 0  [min]& 0.84 (0.18) & 1.57 (0.07) & 1.95 (0.11) & 0.88 (0.06) & 1.73 (0.07) & 1.73 (0.17) & 1.80 (0.08) & 1.51 (0.08) & 2.03 (0.10) \\
Wait time Operator 1 [min] & 0.86 (0.17) & 1.68 (0.12) & 1.85 (0.11) & 0.87 (0.12) & 1.77 (0.06) & 1.71 (0.12) & 1.78 (0.07) & 1.45 (0.08) & 2.06 (0.08) \\
\cmidrule(lr){1-10}
Queue Operator 0 & 2.24 (0.65) & 3.44 (0.84) & 5.29 (0.82) & 1.98 (0.85) & 6.11 (0.57) & 5.84 (0.89) & 8.41 (1.12) & 5.69 (0.73) & 10.4 (1.3) \\
Queue Operator 1 & 2.71 (0.63) & 3.87 (0.52) & 3.23 (0.66) & 2.04 (0.73) & 5.89 (0.83) & 5.53 (1.25) & 7.75 (1.05) & 4.96 (0.67) & 10.8 (1.2) \\
\cmidrule(lr){1-10}
Total Demand & 1112.5 (43.3) & 1519.5 (34.2) & 2154.1 (40.9) & 4746.9 (76.8) & 4990.6 (73.0) & 5682.0 (92.1) & 4683.0 (73.9) & 4311.7 (65.1) & 5231.3 (61.6) \\
Demand Operator 0 & 551.9 (20.2) & 721.3 (25.2) & 1192.4 (32.5) & 2379.4 (54.5) & 2544.8 (49.7) & 2859.9 (51.2) & 2336.2 (56.0) & 2145.4 (36.7) & 2581.1 (59.5) \\
Demand Operator 1 & 560.6 (30.9) & 798.2 (14.5) & 961.7 (21.3) & 2367.5 (36.6) & 2445.8 (29.0) & 2822.1 (44.0) & 2346.8 (39.4) & 2166.3 (34.7) & 2650.2 (42.1) \\
\bottomrule
\end{tabular}%
}
\end{table*}

\subsubsection{Monopolistic vs. Competitive Comparison}
Comparing the best-performing policies across market structures reveals profit losses induced by competition. In San Francisco, the monopolistic joint control achieves 12,286\$ while the best dual-operator configuration yields 10,415\$, a 15.2\% reduction. Washington DC shows a 7.1\% reduction. In NYC Manhattan South, the dual-operator pricing strategy achieves a profit, slightly above the monopolistic joint control. This may reflect the effect that competitive pricing stimulates additional demand in this high-density environment, though the difference (0.7\%) is neglectable. 
Generally, pricing competition drives down the prices compared to the monopolistic setting in all scenarios with up to 27\% in San Francisco. The magnitude of competition-induced profit losses appears related to demand variability. High-CV environments suffer larger losses under competition, as fragmented fleet management amplifies the difficulty of matching supply to volatile demand. In the low-CV NYC scenario, the primary competitive dimension shifts from rebalancing efficiency to price-based market share competition, where fragmentation is less costly.

\subsection{Competitive Policy Analysis in NYC Manhattan South}
\begin{figure}[H]
  \centering
  \subfloat[Operator 0]{
    \includegraphics[width=0.35\columnwidth]{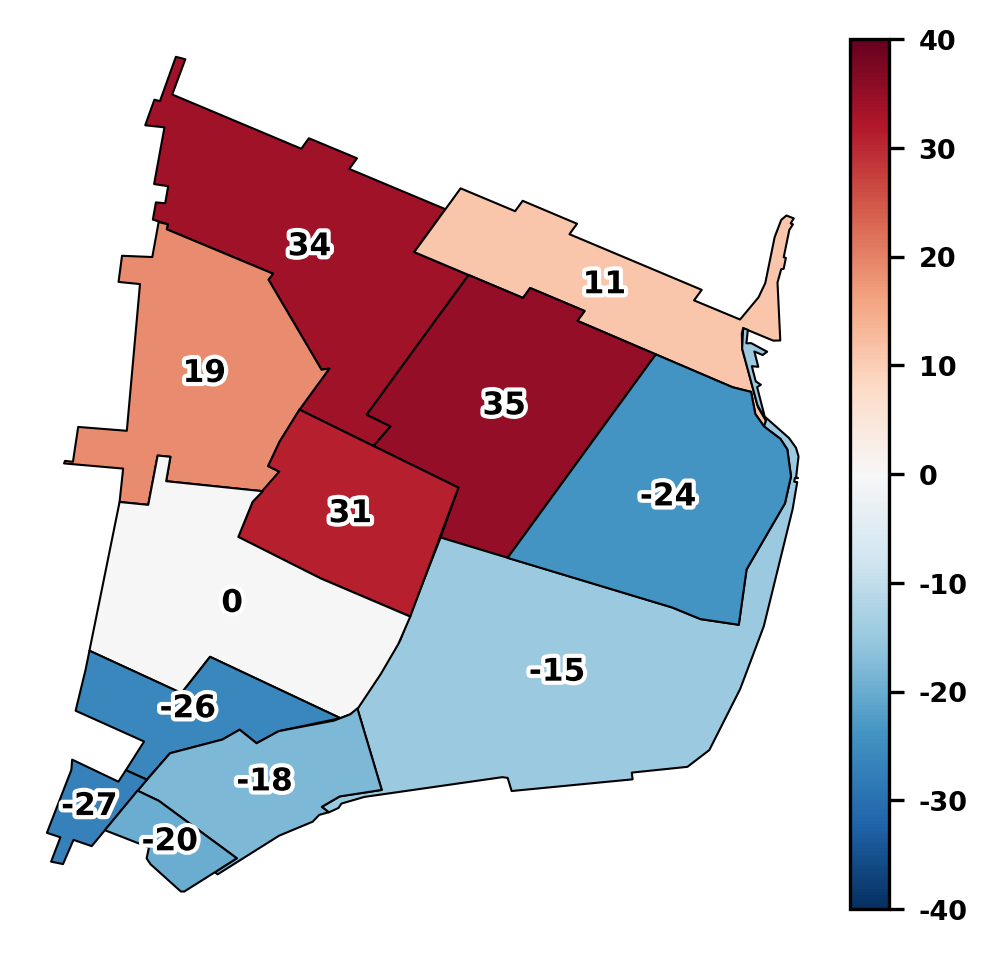}
    \label{fig:reb-dep-a0}
  }\hfill
  \subfloat[Operator 1]{
    \includegraphics[width=0.35\columnwidth]{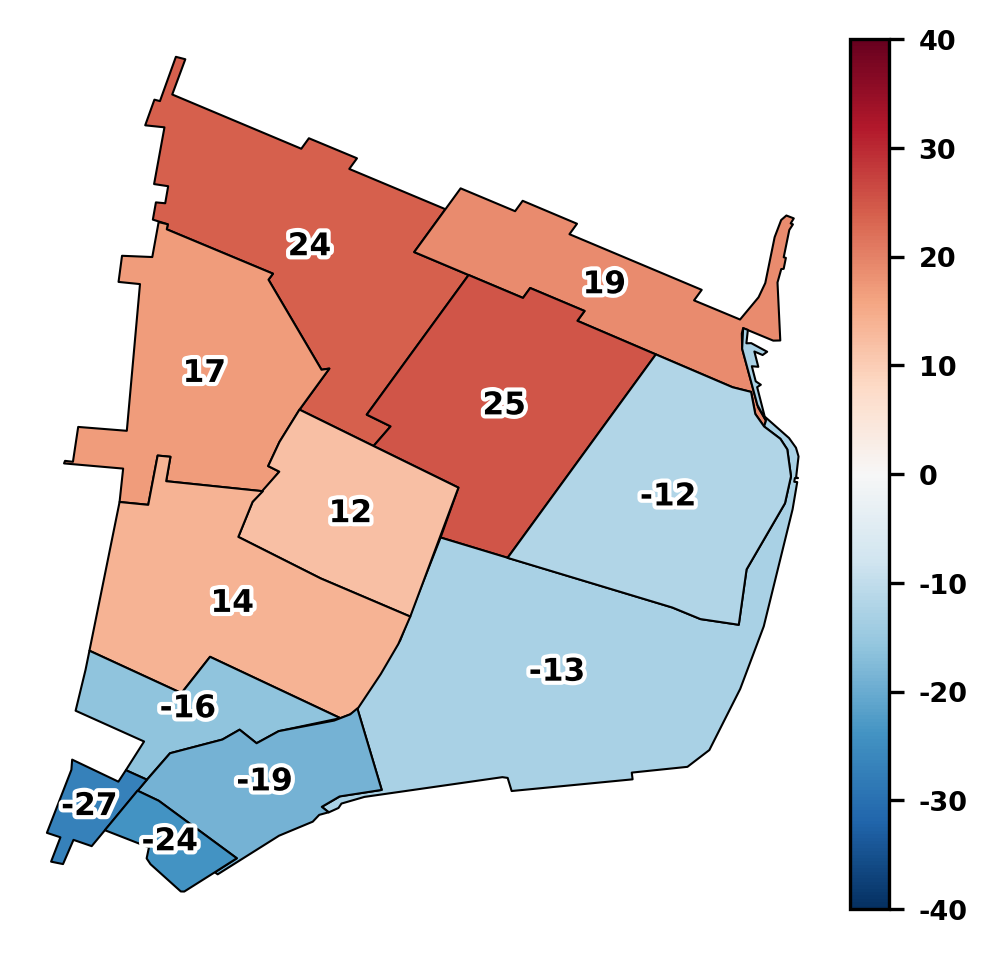}
    \label{fig:reb-dep-a1}
  }
  \caption{Net rebalancing flows for the joint control policy, showing cumulative vehicle movements across all time steps. Red = net receiver, blue = net sender.}
  \label{fig:reb-dep}
\end{figure}

Results in Table~\ref{tab:dual_agent_results} do not directly reveal any specific pricing behavior, as neither operator consistently, strongly undercuts the other on average. However, a granular analysis of the pricing-only policy in NYC-South reveals that operators have learned to employ strategic, slight undercutting in specific areas to capture demand while avoiding ineffective profit losses from excessive undercuts. This is shown in Figure \ref{fig:pricing_mode1_initial}. At $t=0$, Operator 1 selectively undercuts Operator 0 in four high-demand regions, while Operator 0 undercuts in the remaining zones. The strategies are further marked by a strong regional variance of up to a 83\% difference in price scalars between the high-demand regions in the north to low-demand regions in the south. Regarding rebalancing strategies, Figure~\ref{fig:reb-dep} illustrates that rebalancing patterns of both operators closely track demand patterns, with Operator~0 being more targeted in their rebalancing, and Operator~1 choosing a more balanced strategy. 

\begin{figure}[H]
  \centering
  \subfloat[Operator 0]{
    \includegraphics[width=0.35\columnwidth]{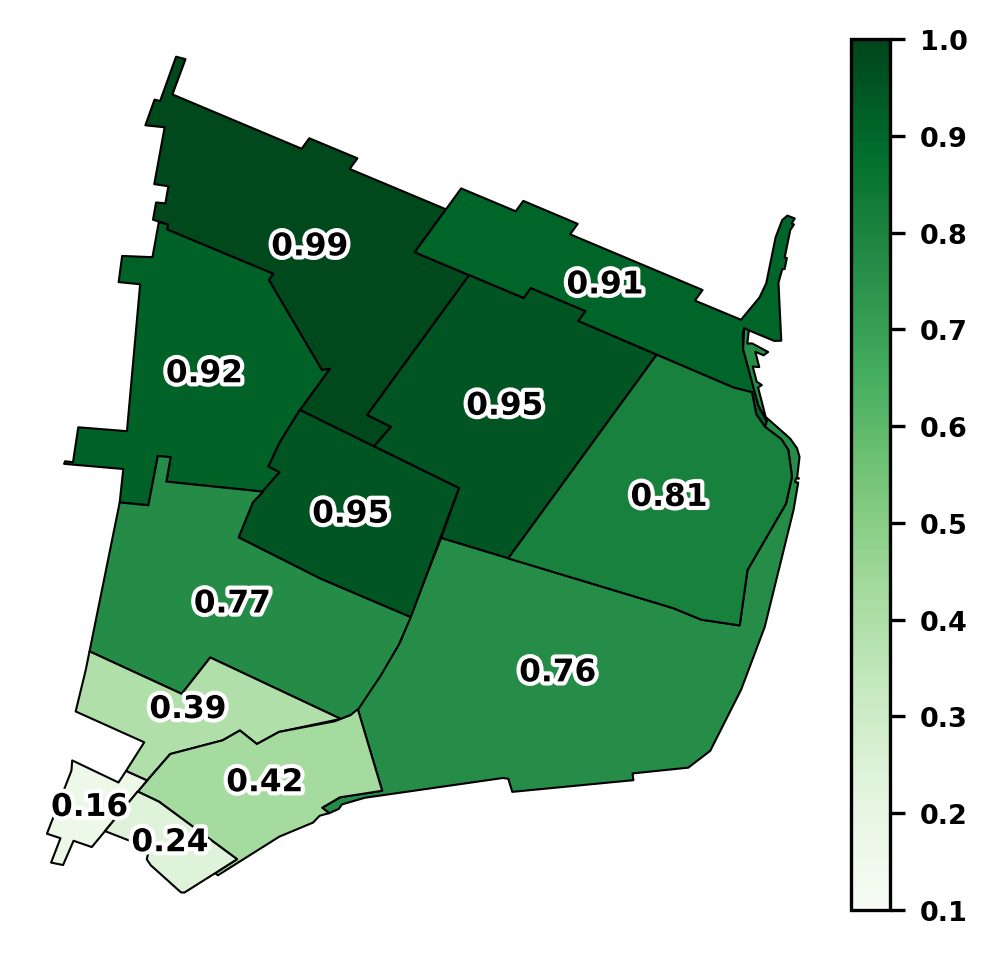}
  }\hfill
  \subfloat[Operator 1]{
    \includegraphics[width=0.35\columnwidth]{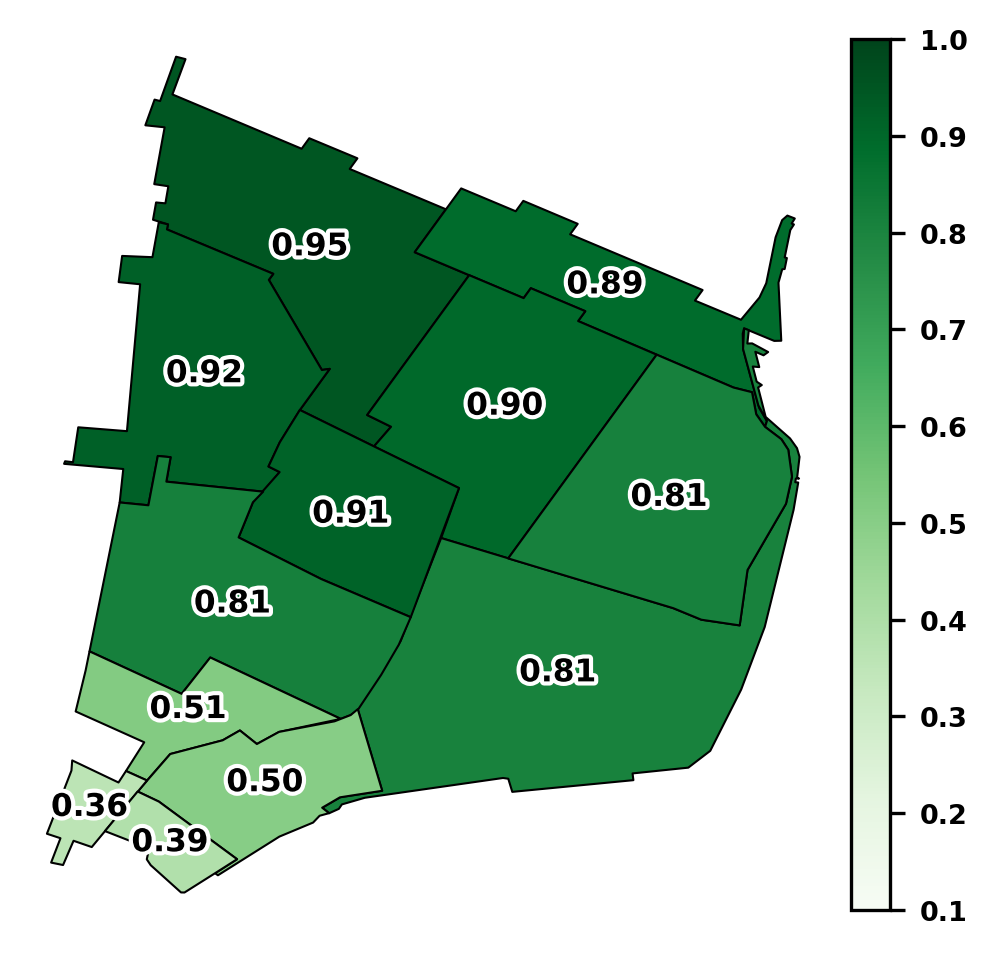}
  }
  \caption{Initial pricing scalars at timestep 0 for the pricing-only policy in NYC Man. South.}
  \label{fig:pricing_mode1_initial}
\end{figure}
\vspace{-0.5cm}
\subsection{Impact of Pricing Information}
We compare scenarios where operators can versus cannot observe competitor prices. Our experiments show that system performance is robust to this information condition across all policy modes, with converged rewards nearly identical, 
suggesting competitor prices may introduce noise rather than actionable signal in the scenarios of simultaneous learning from scratch. Full training curves in appendix.

\subsection{Fleet Size Sensitivity Analysis}
We compare system performance under varying fleet sizes (evenly split between operators) for the joint policy, no control, and uniform distribution baselines in NYC Man. South. The joint policy dynamically lowers price scalars as fleet size grows (from 1.04 to 0.71) to maintain utilization of the larger fleet, serving up to 64\% more passengers than NC. Total reward peaks at $1{,}050$ vehicles and declines beyond this point as rebalancing costs outpace revenue gains, suggesting diminishing profit from fleet expansion. Full results are available appendix.

\subsection{Fleet Split Sensitivity Analysis}
We evaluate the joint policy in NYC Man. South under five asymmetric fleet splits ranging from 5:5 to 1:9 between the two operators. Total reward is highest at the 4:6 split (17{,}567 \$) and degrades at more extreme asymmetries, while total served demand remains stable across configurations (3{,}500-3{,}577). The smaller operator compensates for its limited capacity by raising prices (from 0.96 at 5:5 to 1.10 at 1:9), whereas the larger fleet of the other operator allows to consistently undercut (0.93-0.96). Theese results indicate, that fleet size imbalances reduces direct price competition. Full results are available in appendix. 

\subsection{Regional Wage Heterogeneity}
Lastly, we evaluate the impact of spatially varying passenger incomes by conducting experiments in NYC Manhattan South, where regional wages range from \$10 to \$30, with a total fleet of 650 vehicles evenly split between operators. Notably, rebalancing flows reveal spatial adaptation, with both operators repositioning vehicles from low-income regions toward high-income areas where demand is concentrated, while pricing scalars are increased to exploit higher willingness to pay-  full results are provided in appendix.

\section{Summary and Conclusion}
\label{sec:conclusion}
This paper introduced a competitive multi-operator reinforcement learning framework for joint pricing and fleet rebalancing in AMoD systems to study how competition impacts operator policy learning.  By embedding a discrete choice model within the learning loop, passenger allocation emerges endogenously through utility maximization, coupling pricing actions to demand competition. Crucially, we empirically show that learning-based approaches remain robust to the additional stochasticity introduced by competition. Competitive agents converge to stable rewards and learn policies that effectively rebalance fleets and set dynamic prices while accounting for unobserved competitor behavior.

Experiments across three cities yield several key findings. Under monopoly, joint pricing and rebalancing consistently maximizes profit. Competition fundamentally alters which strategies dominate: while joint control excels under monopoly, competitive markets can favor specialized policies. 

Competition drives prices down, benefiting passengers through lower fares, but multi-operator fleet management increases waiting times, illustrating mixed welfare effects. 
Sensitivity analyses show that operators dynamically adapt to fleet size changes, asymmetric fleet splits reduces direct pricing competition where smaller operators raise prices while larger ones undercut, and wage heterogeneity drives fleet repositioning toward high-income areas and increase the prices to exploit higher willingness to pay. 

These findings suggest several directions for future work: incorporating granular waiting time beyond a maximum threshold into passenger utilities to introduce a direct trade-off between profitability and service quality, investigating emergent behaviors induced by asymmetric architectures or learning parameters, and explicitly modeling collusive behavior to analyze its impact on pricing and consumer welfare.\\

	\bibliographystyle{IEEEtran}
	\bibliography{root} 

\appendix

\section*{Vehicle Rebalancing Model}
 At each time step $t$, the model determines the rebalancing flows $\{y_{i,j,o}^t\}$ for operator $o$ that minimize the total rebalancing cost while satisfying the desired vehicle distribution specified by the actor network. The optimization problem is formulated as:

\vspace{-0.5em}
\begin{align}
\min \quad & \sum_{(i,j) \in \mathcal{E}} c_{i,j,o}^t y_{i,j,o}^t \label{eq:rebalancing_objective} \\
\text{s.t.} \quad & \sum_{j \neq i} (y_{j,i,o}^t - y_{i,j,o}^t) + m_{i,o}^t \geq \tilde{m}_{i,o}^t, \quad i \in \mathcal{V} \label{eq:rebalancing_flow_balance} \\
& \sum_{j \neq i} y_{i,j,o}^t \leq m_{i,o}^t, \quad i \in \mathcal{V} \label{eq:rebalancing_capacity} \\
& y_{i,j,o}^t \geq 0, \quad (i,j) \in \mathcal{E} \label{eq:rebalancing_nonnegative}
\end{align}
where $\tilde{m}_{i,o}^t$ denotes the number of desired vehicles at region $i$ for operator $o$ at time $t$. Objective~\eqref{eq:rebalancing_objective} minimizes the rebalancing cost. Constraint~\eqref{eq:rebalancing_flow_balance} ensures that the desired vehicle number is satisfied, accounting for the current idle vehicles $m_{i,o}^t$ and the net inflow of rebalanced vehicles. Constraint~\eqref{eq:rebalancing_capacity} limits the rebalancing flow from each region by the number of available idle vehicles. The desired vehicle distribution is calculated by $\tilde{m}_{i,o}^t = \lfloor w_{i,o}^t \sum_{i \in \mathcal{V}} m_{i,o}^t \rfloor$, where $w_{i,o}^t$ is the rebalancing weight output by the actor network for region $i$. Note that the constraint matrix of this network flow problem is totally unimodular, and since both $m_{i,o}^t$ and $\tilde{m}_{i,o}^t$ are integer-valued, the optimal solution is guaranteed to be integral.
\section*{Parameters Used in Training}
\begin{table}[H]
\centering
\label{tab:hyperparameters}
\begin{tabular}{llc}
\toprule
\textbf{Category} & \textbf{Hyperparameter} & \textbf{Value} \\
\midrule
\multirow{8}{*}{\textit{Training}} 
 & Actor learning rate ($\alpha_\pi$) & $2 \times 10^{-4}$ \\
 & Critic learning rate ($\alpha_V$) & $4 \times 10^{-4}$ \\
 & Discount factor ($\gamma$) & 0.97 \\
 & Reward scaling factor & 4,000 \\
 & Actor gradient clip & 1,000 \\
 & Critic gradient clip & 1,000 \\
 & Critic warmup episodes & 50 \\
 & Training episodes & 150{,}000 \\
\midrule
\multirow{3}{*}{\textit{Network}} 
 & Hidden layer size & 256 \\
 & Look-ahead horizon ($T$) & 6 \\
 & Scale factor & 0.01 \\
\midrule
\multirow{2}{*}{\textit{Price Scalars}} 
 & Observe OD-prices & Yes \\
 & OD-price scalars & No \\
\bottomrule
\end{tabular}
\end{table}

\section*{Choice Model Sensitivity}
\begin{figure}[H]
  \centering
  \includegraphics[width=0.5\textwidth]{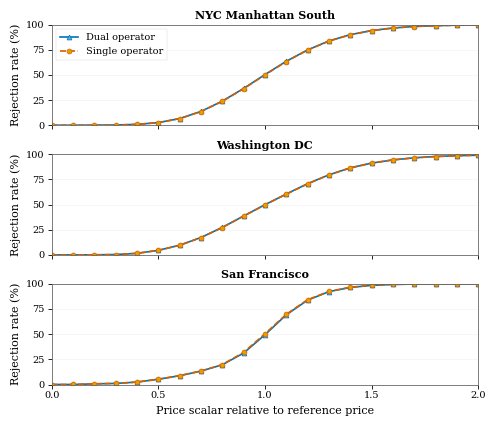}
  \caption{Rejection rate versus price scalar relative to the historical reference price across studied datasets for both single and dual-operator setups, with the model calibrated to a 50\% rejection rate at the historical reference price.}
  \label{fig:price_sensitivity}
\end{figure}

\twocolumn[{%
  \centering
  \section*{Three-Step Control Architecture}
  \includegraphics[width=1.0\textwidth]{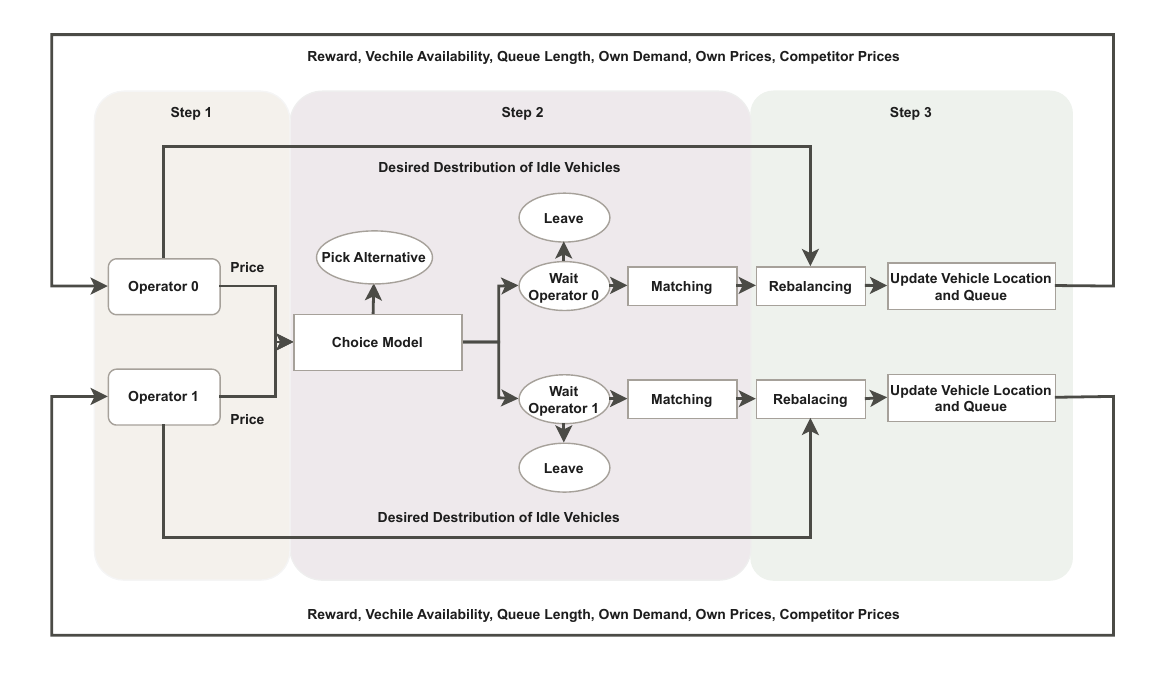}
  \captionof{figure}{Three-step control architecture for dual-operator AMoD control. Step 1: operators formulate pricing and desired idle-vehicle distribution policies. Step 2: passenger assignment via choice model, queueing, and matching. Step 3: idle-vehicle rebalancing and update of vehicle positions and queues.}
  \label{fig:dual_agent_flow}
  \vspace{1cm}
  \section*{Operator Architecture}
  \includegraphics[width=1.0\textwidth]{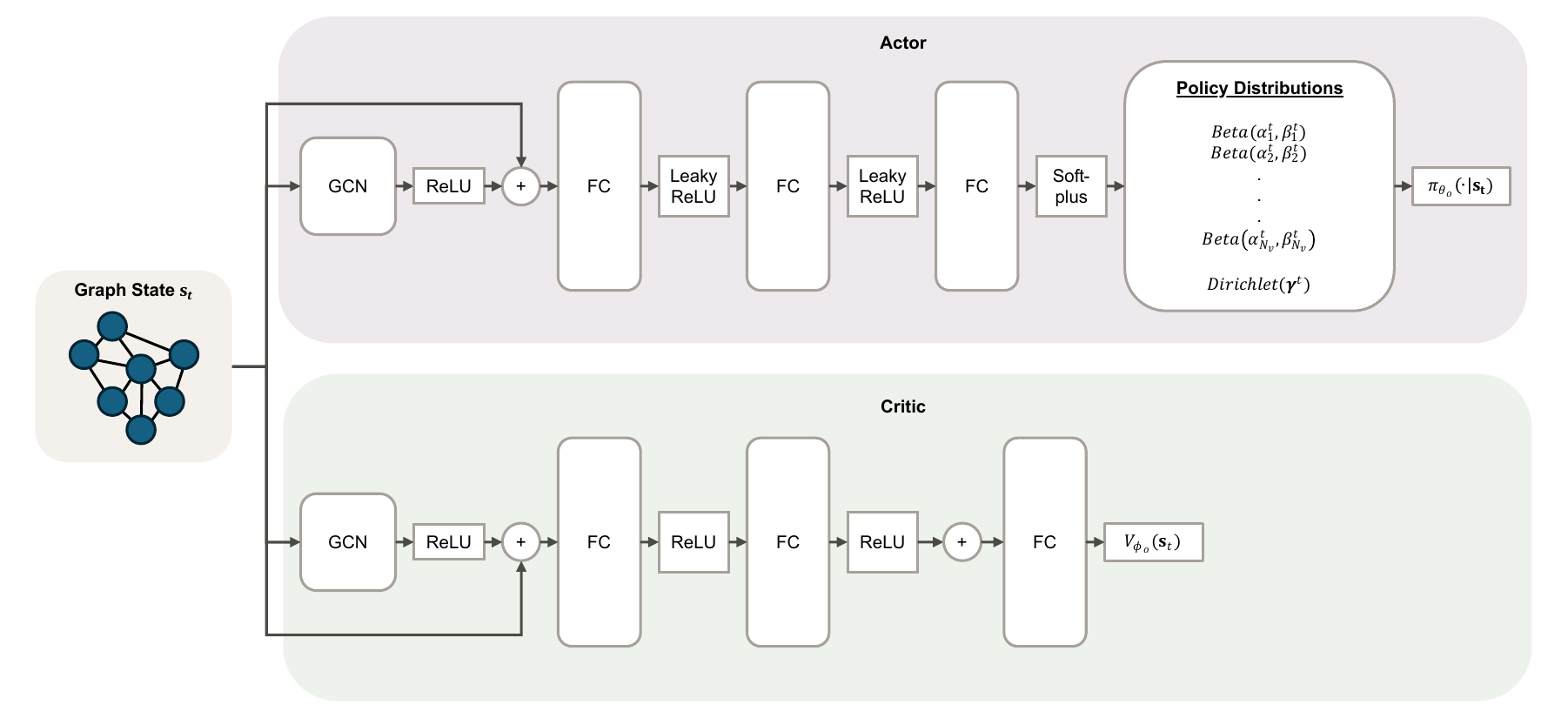}
  \captionof{figure}{The Actor-Critic architecture employed by the operators. Each operator maintains independent actor and critic networks.}
  \label{fig:arch}
}]

\section*{Pricing Scalars under Pricing-only Policy NYC}
\begin{figure}[H]
  \centering
  \subfloat[Operator 0]{
    \includegraphics[width=0.45\columnwidth]{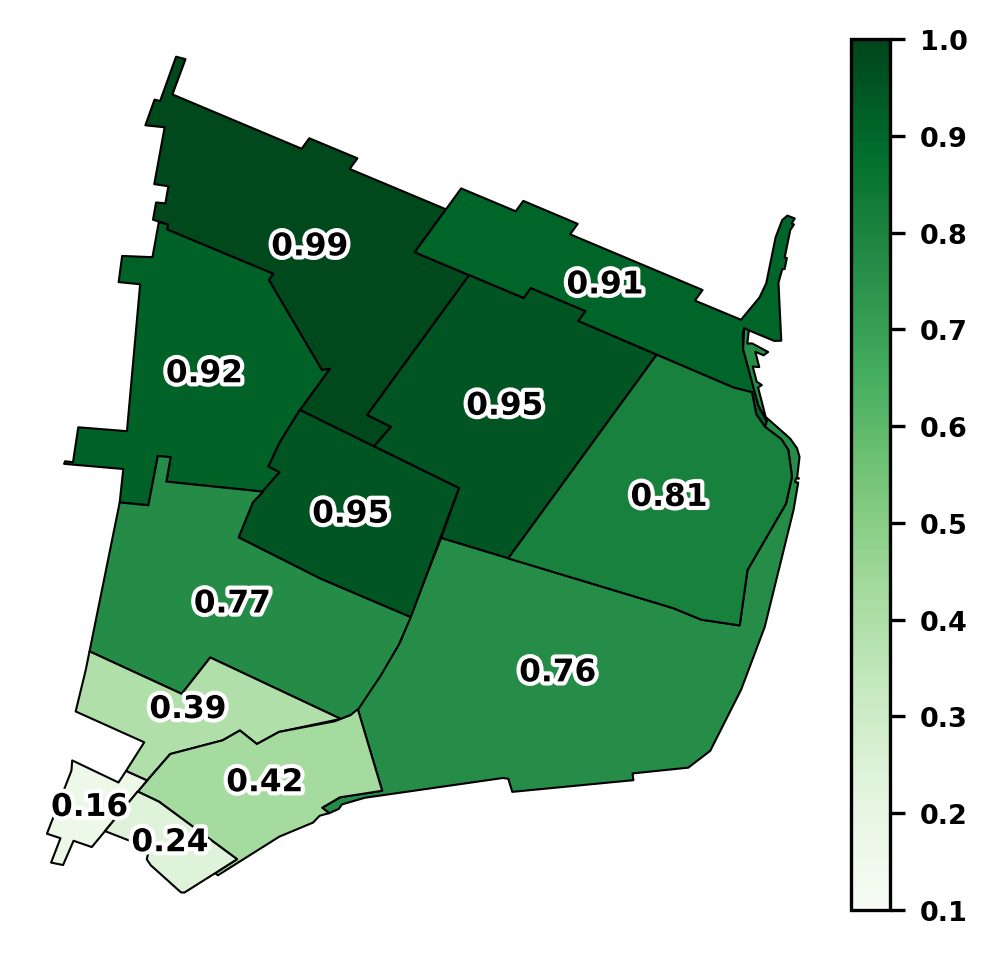}
  }\hfill
  \subfloat[Operator 1]{
    \includegraphics[width=0.45\columnwidth]{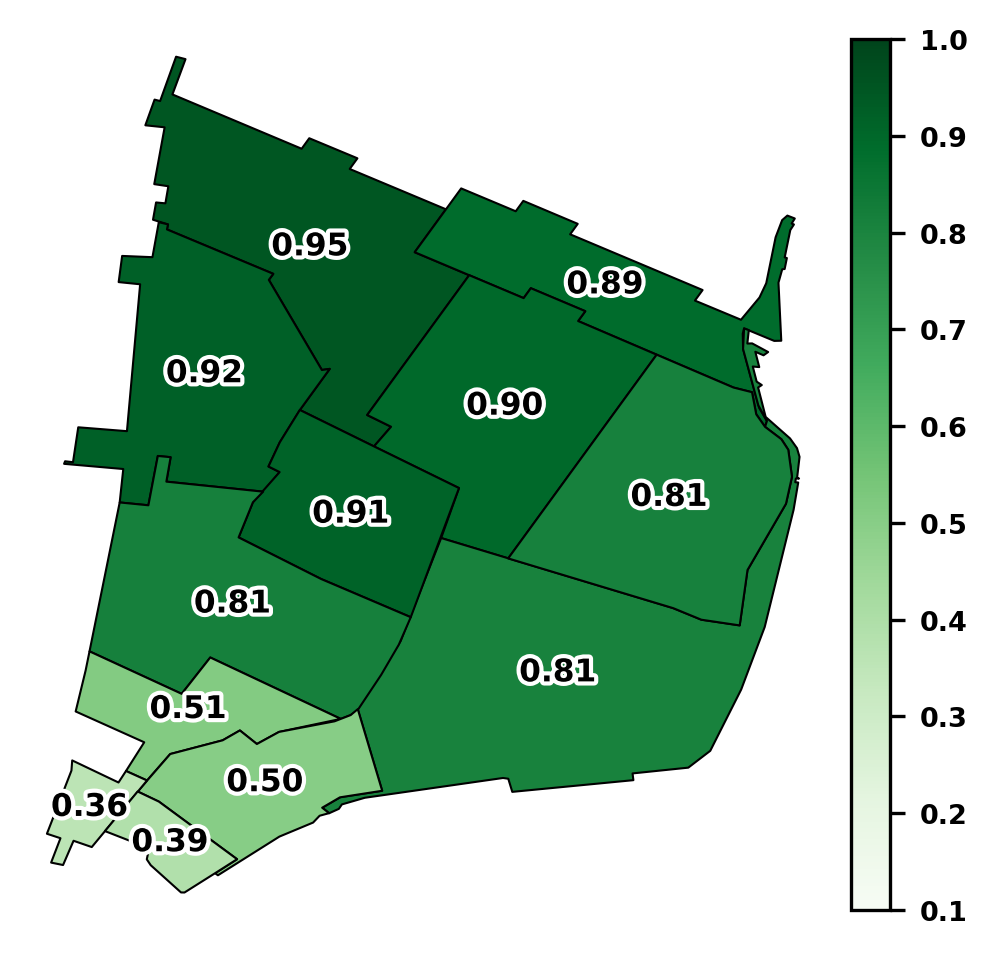}
  }
  \caption{Initial pricing policies at timestep 0 for the pricing-only policy.}
  \label{fig:pricing_mode1_initial}
\end{figure}

\begin{figure}[H]
  \centering
  \subfloat[Operator 0]{
    \includegraphics[width=0.45\columnwidth]{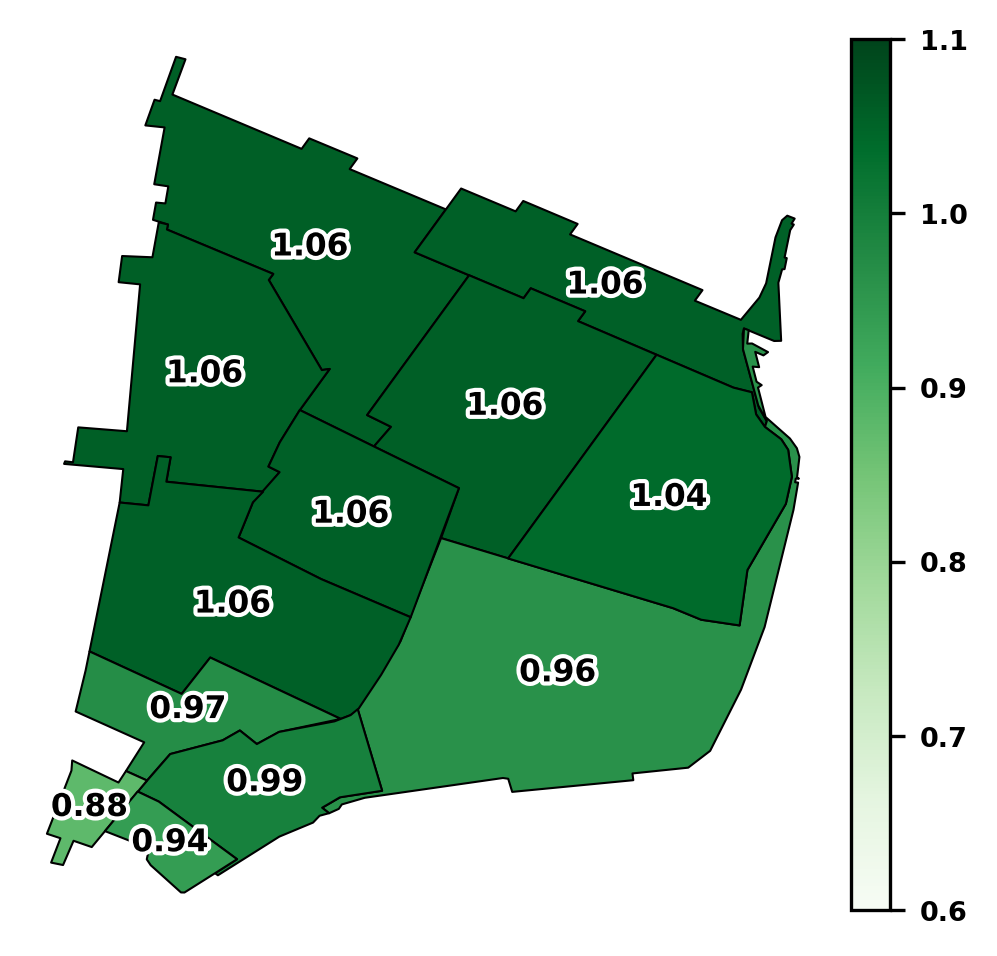}
  }\hfill
  \subfloat[Operator 1]{
    \includegraphics[width=0.45\columnwidth]{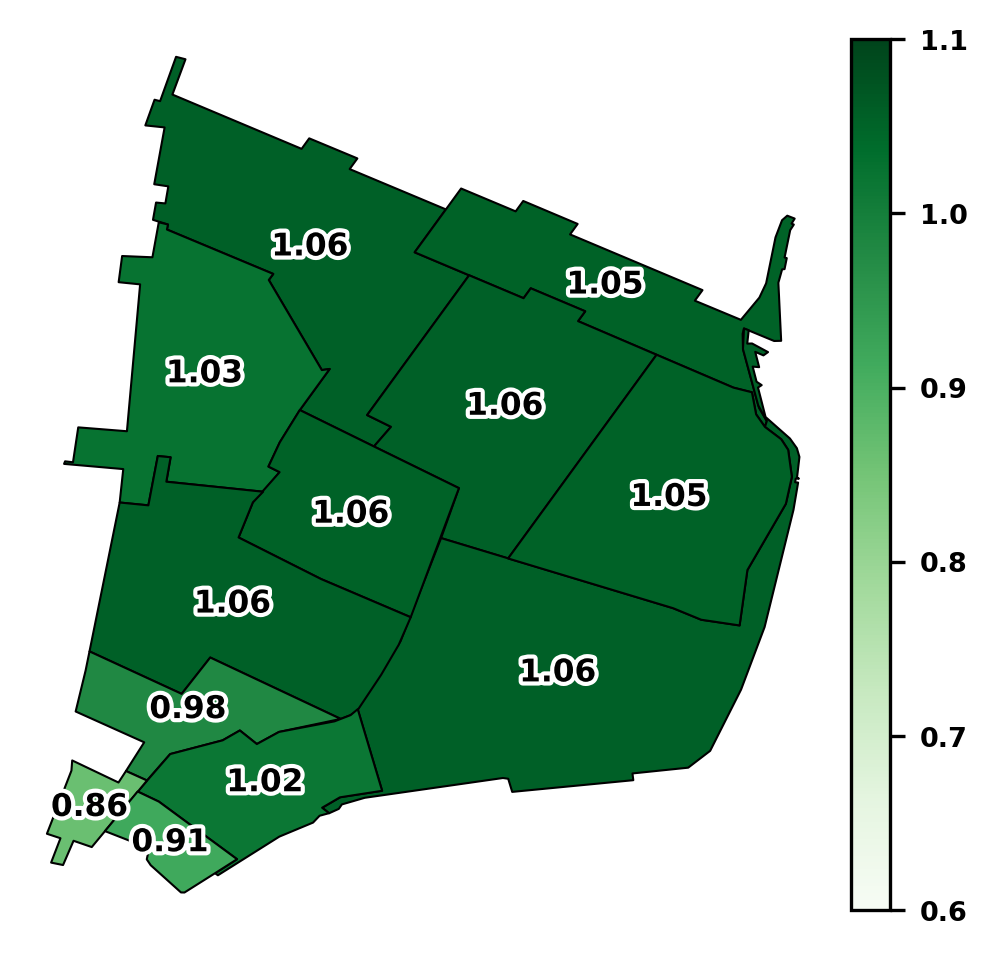}
  }
  \caption{Final pricing policies at time step 19 for the pricing-only policy.}
  \label{fig:pricing_mode1_final}
\end{figure}

\begin{figure}[H]
  \centering
  \subfloat[Operator 0]{
    \includegraphics[width=0.45\columnwidth]{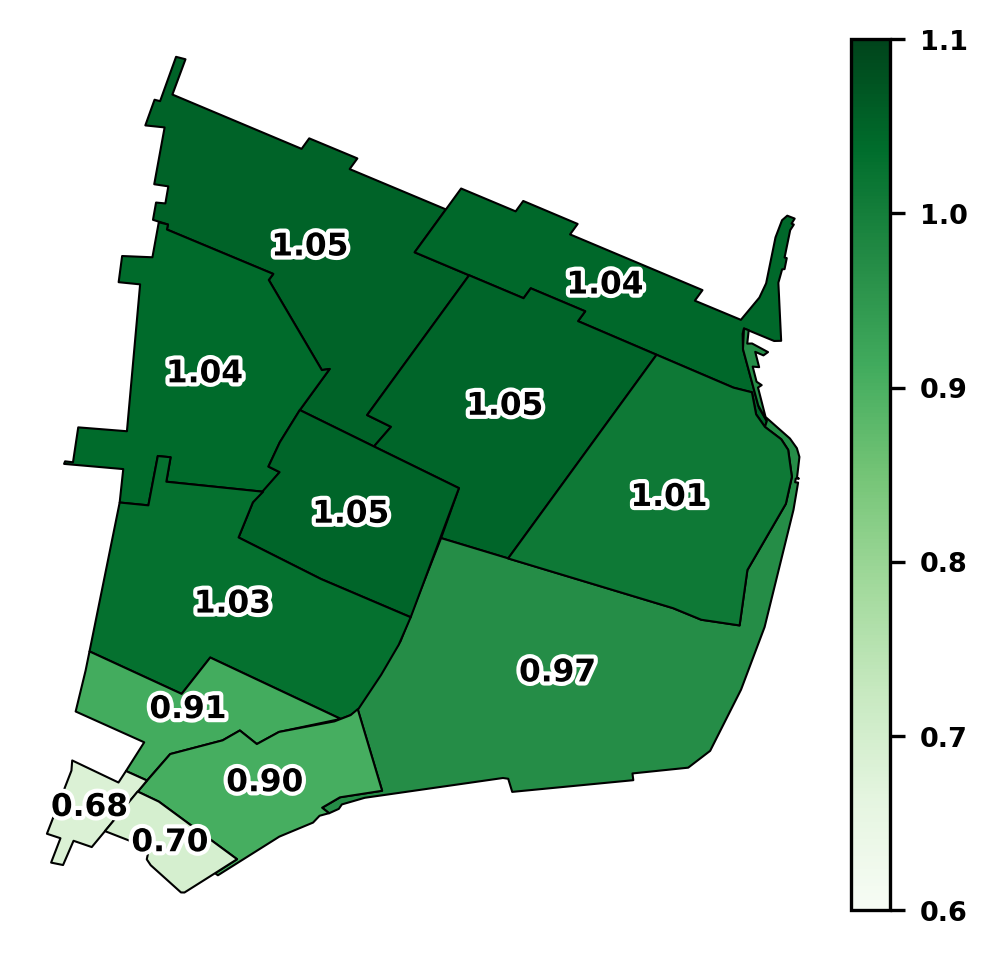}
  }\hfill
  \subfloat[Operator 1]{
    \includegraphics[width=0.45\columnwidth]{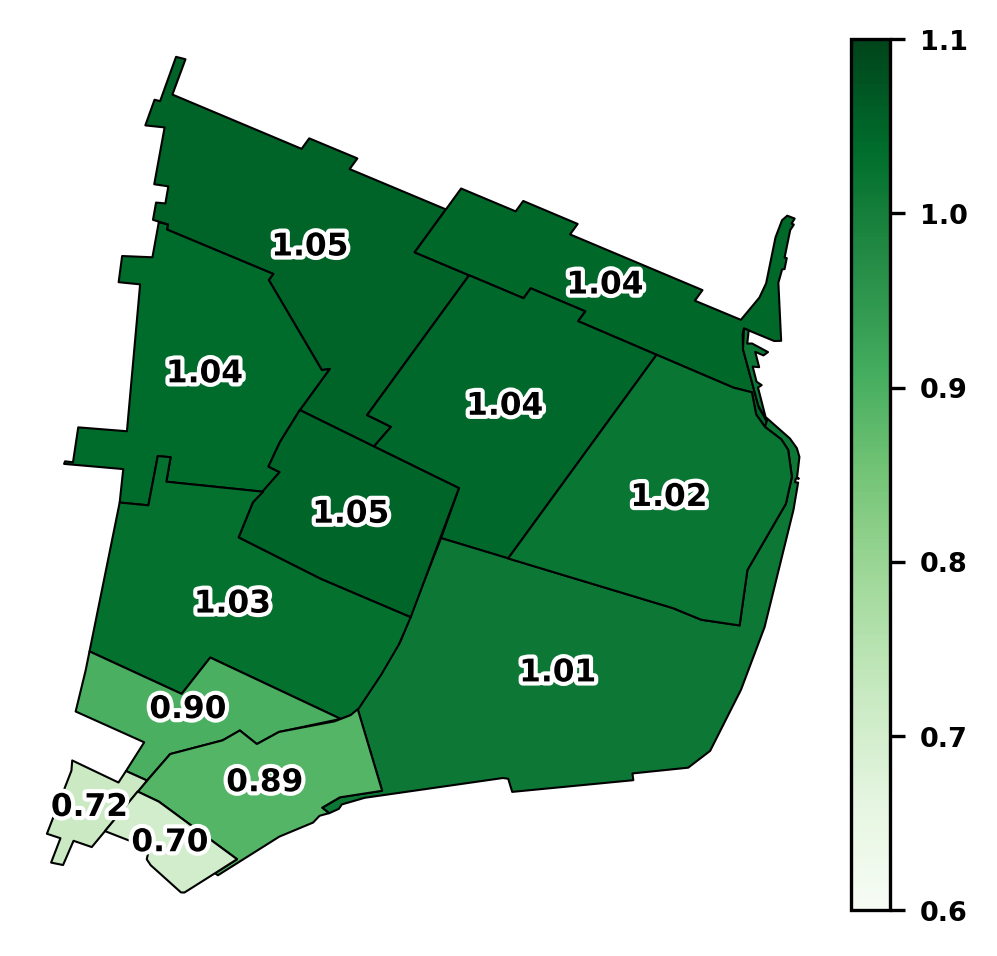}
  }
  \caption{Time-averaged pricing scalars for the pricing-only policy, computed across all 20 time steps.}
  \label{fig:pricing_mode1_average}
\end{figure}

\twocolumn[{%
  \centering
  \section*{Fleet Size Sensitivity}
  \vspace{0.3cm}
  \resizebox{\textwidth}{!}{%
  \begin{tabular}{l cccccc ccc ccccc}
  \toprule
  & \multicolumn{6}{c}{\textbf{Joint Policy}} & \multicolumn{3}{c}{\textbf{NC}} & \multicolumn{5}{c}{\textbf{UD}} \\
  \cmidrule(lr){2-7} \cmidrule(lr){8-10} \cmidrule(lr){11-15}
  Fleet & Reward & Served & Rebal. & Rebal. & Price & Price & Reward & Served & Price & Reward & Served & Rebal. & Rebal. & Price \\
  Size & & & Costs & Trips & A0 & A1 & & & Both & & & Costs & Trips & Both \\
  \midrule
  450 & \makecell{14170.5 \\ (251.7)} & \makecell{2563.9 \\ (28.4)} & \makecell{865.6 \\ (66.9)} & \makecell{177.3 \\ (16.0)} & \makecell{1.04 \\ (0.00)} & \makecell{1.04 \\ (0.00)} & \makecell{12270.7 \\ (251.8)} & \makecell{2550.6 \\ (15.9)} & \makecell{1.00 \\ (0.00)} & \makecell{13217.4 \\ (153.7)} & \makecell{2550.6 \\ (15.9)} & \makecell{989.0 \\ (70.0)} & \makecell{200.6 \\ (16.6)} & \makecell{1.00 \\ (0.00)} \\
  \midrule
  650 & \makecell{17481.1 \\ (258.7)} & \makecell{3561.1 \\ (29.0)} & \makecell{1379.0 \\ (83.7)} & \makecell{284.7 \\ (19.0)} & \makecell{0.96 \\ (0.00)} & \makecell{0.96 \\ (0.00)} & \makecell{16048.4 \\ (416.6)} & \makecell{2894.2 \\ (73.9)} & \makecell{1.00 \\ (0.00)} & \makecell{17652.0 \\ (277.5)} & \makecell{3496.5 \\ (32.2)} & \makecell{1865.1 \\ (98.3)} & \makecell{387.9 \\ (19.6)} & \makecell{1.00 \\ (0.00)} \\
  \midrule
  850 & \makecell{19513.1 \\ (288.1)} & \makecell{4538.7 \\ (38.9)} & \makecell{1871.0 \\ (79.9)} & \makecell{386.3 \\ (16.7)} & \makecell{0.88 \\ (0.00)} & \makecell{0.88 \\ (0.00)} & \makecell{18683.9 \\ (461.8)} & \makecell{3367.4 \\ (83.3)} & \makecell{1.00 \\ (0.00)} & \makecell{19941.6 \\ (334.8)} & \makecell{4171.5 \\ (44.7)} & \makecell{3401.8 \\ (128.6)} & \makecell{761.5 \\ (27.2)} & \makecell{1.00 \\ (0.00)} \\
  \midrule
  1050 & \makecell{20373.0 \\ (198.7)} & \makecell{5512.9 \\ (30.1)} & \makecell{2152.1 \\ (77.1)} & \makecell{446.8 \\ (19.7)} & \makecell{0.79 \\ (0.00)} & \makecell{0.79 \\ (0.00)} & \makecell{20667.8 \\ (467.6)} & \makecell{3714.9 \\ (83.3)} & \makecell{1.00 \\ (0.00)} & \makecell{20450.0 \\ (543.4)} & \makecell{4538.1 \\ (72.0)} & \makecell{5016.9 \\ (155.3)} & \makecell{1187.7 \\ (41.2)} & \makecell{1.00 \\ (0.00)} \\
  \midrule
  1250 & \makecell{20305.6 \\ (235.2)} & \makecell{6440.1 \\ (32.2)} & \makecell{2509.8 \\ (117.9)} & \makecell{526.9 \\ (29.9)} & \makecell{0.71 \\ (0.00)} & \makecell{0.70 \\ (0.00)} & \makecell{21955.0 \\ (397.2)} & \makecell{3934.1 \\ (71.2)} & \makecell{1.00 \\ (0.00)} & \makecell{19798.1 \\ (526.1)} & \makecell{4639.2 \\ (77.2)} & \makecell{6259.8 \\ (192.8)} & \makecell{1501.7 \\ (46.1)} & \makecell{1.00 \\ (0.00)} \\
  \bottomrule
  \end{tabular}
  }
  \label{tab:fleet_sensitivity}

  \vspace{0.8cm}

  \section*{Fleet Split Sensitivity}
  \vspace{0.3cm}
  \resizebox{\textwidth}{!}{%
  \begin{tabular}{c c c c c c c c c c c c}
  \toprule
  \makecell{Fleet Split \\ (O0:O1)} & \makecell{Total \\ Reward} & \makecell{O0 \\ Reward} & \makecell{O1 \\ Reward} & \makecell{Total Rebal. \\ Trips} & \makecell{O0 Rebal. \\ Trips} & \makecell{O1 Rebal. \\ Trips} & \makecell{O0 \\ Price} & \makecell{O1 \\ Price} & \makecell{Total \\ Served} & \makecell{O0 \\ Served} & \makecell{O1 \\ Served} \\
  \midrule
  5:5 & \makecell{17481.1 \\ (258.7)} & \makecell{8717.0 \\ (249.1)} & \makecell{8764.1 \\ (223.5)} & \makecell{284.7 \\ (19.0)} & \makecell{144.9 \\ (17.8)} & \makecell{139.8 \\ (14.3)} & \makecell{0.96 \\ (0.00)} & \makecell{0.96 \\ (0.00)} & \makecell{3561.1 \\ (29.0)} & \makecell{1773.9 \\ (26.1)} & \makecell{1787.2 \\ (24.0)} \\
  \midrule
  4:6 & \makecell{17567.0 \\ (231.0)} & \makecell{7295.2 \\ (136.2)} & \makecell{10271.7 \\ (243.7)} & \makecell{279.2 \\ (17.8)} & \makecell{116.1 \\ (12.6)} & \makecell{163.1 \\ (14.6)} & \makecell{0.98 \\ (0.00)} & \makecell{0.95 \\ (0.00)} & \makecell{3577.2 \\ (27.9)} & \makecell{1437.2 \\ (14.7)} & \makecell{2140.0 \\ (30.1)} \\
  \midrule
  3:7 & \makecell{17434.2 \\ (241.7)} & \makecell{5772.6 \\ (143.7)} & \makecell{11661.6 \\ (259.2)} & \makecell{285.3 \\ (19.8)} & \makecell{79.6 \\ (11.5)} & \makecell{205.7 \\ (18.9)} & \makecell{1.01 \\ (0.00)} & \makecell{0.94 \\ (0.00)} & \makecell{3562.5 \\ (32.1)} & \makecell{1090.1 \\ (17.4)} & \makecell{2472.4 \\ (33.0)} \\
  \midrule
  2:8 & \makecell{16968.2 \\ (263.3)} & \makecell{4140.0 \\ (88.8)} & \makecell{12828.2 \\ (244.2)} & \makecell{294.5 \\ (21.2)} & \makecell{41.3 \\ (4.9)} & \makecell{253.2 \\ (19.8)} & \makecell{1.04 \\ (0.00)} & \makecell{0.93 \\ (0.00)} & \makecell{3530.7 \\ (33.5)} & \makecell{742.3 \\ (13.6)} & \makecell{2788.4 \\ (31.9)} \\
  \midrule
  1:9 & \makecell{16655.4 \\ (291.8)} & \makecell{2348.7 \\ (77.9)} & \makecell{14306.7 \\ (299.7)} & \makecell{311.6 \\ (22.3)} & \makecell{14.1 \\ (3.0)} & \makecell{297.5 \\ (22.9)} & \makecell{1.10 \\ (0.00)} & \makecell{0.93 \\ (0.00)} & \makecell{3499.9 \\ (35.8)} & \makecell{380.7 \\ (8.7)} & \makecell{3119.2 \\ (38.1)} \\
  \bottomrule
  \end{tabular}
  }
  \label{tab:fleet_split_results_dual}
}]

\clearpage

\twocolumn[{%
  \centering
  \section*{Impact of Pricing Information}
  \vspace{0.3cm}
  \resizebox{\textwidth}{!}{%
  \begin{tabular}{l ccc ccc}
  \toprule
  & \multicolumn{3}{c}{\textbf{No Information Sharing}} & \multicolumn{3}{c}{\textbf{Information Sharing}} \\
  \cmidrule(lr){2-4} \cmidrule(lr){5-7}
  Policy & Reb. & Pricing & Joint & Reb. & Pricing & Joint \\
  \midrule
  Total Reward & 18096.6 (296.6) & 18983.6 (267.4) & 17689.4 (242.4) & 18133.7 (330.8) & 18631.6 (376.7) & 17481.1 (258.7) \\
  Reward Operator 0 & 8981.1 (227.9) & 9413.0 (111.9) & 8951.6 (212.4) & 9010.8 (242.6) & 9351.6 (161.4) & 8717.0 (249.1) \\
  Reward Operator 1 & 9115.5 (317.0) & 9570.6 (192.9) & 8737.8 (190.5) & 9123.0 (341.4) & 9280.1 (265.1) & 8764.1 (223.5) \\
  \cmidrule(lr){1-7}
  Total Rebalancing Costs & 1544.4 (110.7) & --- & 1349.7 (76.5) & 1600.2 (95.4) & --- & 1379.0 (83.7) \\
  Rebal.\ Costs Operator 0 & 806.0 (78.6) & --- & 662.5 (65.8) & 813.5 (73.6) & --- & 699.6 (76.7) \\
  Rebal.\ Costs Operator 1 & 738.5 (103.5) & --- & 687.1 (59.2) & 786.8 (107.0) & --- & 679.4 (63.1) \\
  \cmidrule(lr){1-7}
  Total Rebalance Trips & 327.8 (25.4) & --- & 276.5 (16.8) & 338.4 (21.2) & --- & 284.7 (19.0) \\
  Rebal.\ Trips Operator 0 & 172.0 (20.6) & --- & 134.7 (15.3) & 173.5 (18.2) & --- & 144.9 (17.8) \\
  Rebal.\ Trips Operator 1 & 155.8 (23.6) & --- & 141.8 (12.4) & 164.9 (24.2) & --- & 139.8 (14.3) \\
  \cmidrule(lr){1-7}
  Total Served Demand & 3521.9 (34.2) & 3388.9 (38.1) & 3579.5 (29.0) & 3535.8 (39.4) & 3337.6 (52.3) & 3561.1 (29.0) \\
  Served Demand Operator 0 & 1751.8 (26.1) & 1678.1 (13.8) & 1800.4 (25.4) & 1758.1 (29.4) & 1674.3 (19.1) & 1773.9 (26.1) \\
  Served Demand Operator 1 & 1770.1 (36.8) & 1710.8 (28.8) & 1779.1 (21.0) & 1777.7 (39.5) & 1663.3 (38.7) & 1787.2 (24.0) \\
  \cmidrule(lr){1-7}
  Price Operator 0 & --- & 0.96 (0.00) & 0.96 (0.00) & --- & 0.95 (0.00) & 0.96 (0.00) \\
  Price Operator 1 & --- & 0.95 (0.01) & 0.96 (0.00) & --- & 0.95 (0.01) & 0.96 (0.00) \\
  \cmidrule(lr){1-7}
  Wait/mins Operator 0 & 1.82 (0.08) & 1.34 (0.06) & 1.97 (0.10) & 1.80 (0.08) & 1.51 (0.08) & 2.03 (0.10) \\
  Wait/mins Operator 1 & 1.81 (0.07) & 1.37 (0.07) & 2.06 (0.09) & 1.78 (0.07) & 1.45 (0.08) & 2.06 (0.08) \\
  \cmidrule(lr){1-7}
  Queue Operator 0 & 8.37 (1.09) & 4.35 (0.70) & 9.89 (1.12) & 8.41 (1.12) & 5.69 (0.73) & 10.4 (1.3) \\
  Queue Operator 1 & 7.78 (1.21) & 4.62 (0.80) & 10.7 (1.4) & 7.75 (1.05) & 4.96 (0.67) & 10.8 (1.2) \\
  \cmidrule(lr){1-7}
  Total Demand & 4683.0 (73.9) & 4202.3 (55.2) & 5185.7 (58.2) & 4683.0 (73.9) & 4311.7 (65.1) & 5231.3 (61.6) \\
  Demand Operator 0 & 2336.2 (56.0) & 2075.3 (34.3) & 2552.0 (52.5) & 2336.2 (56.0) & 2145.4 (36.7) & 2581.1 (59.5) \\
  Demand Operator 1 & 2346.8 (39.4) & 2127.0 (31.8) & 2633.7 (34.4) & 2346.8 (39.4) & 2166.3 (34.7) & 2650.2 (42.1) \\
  \bottomrule
  \end{tabular}
  }
  \label{tab:info_sharing_comparison}
}]
\begin{figure}[H]
\centering
\includegraphics[width=0.55\columnwidth]{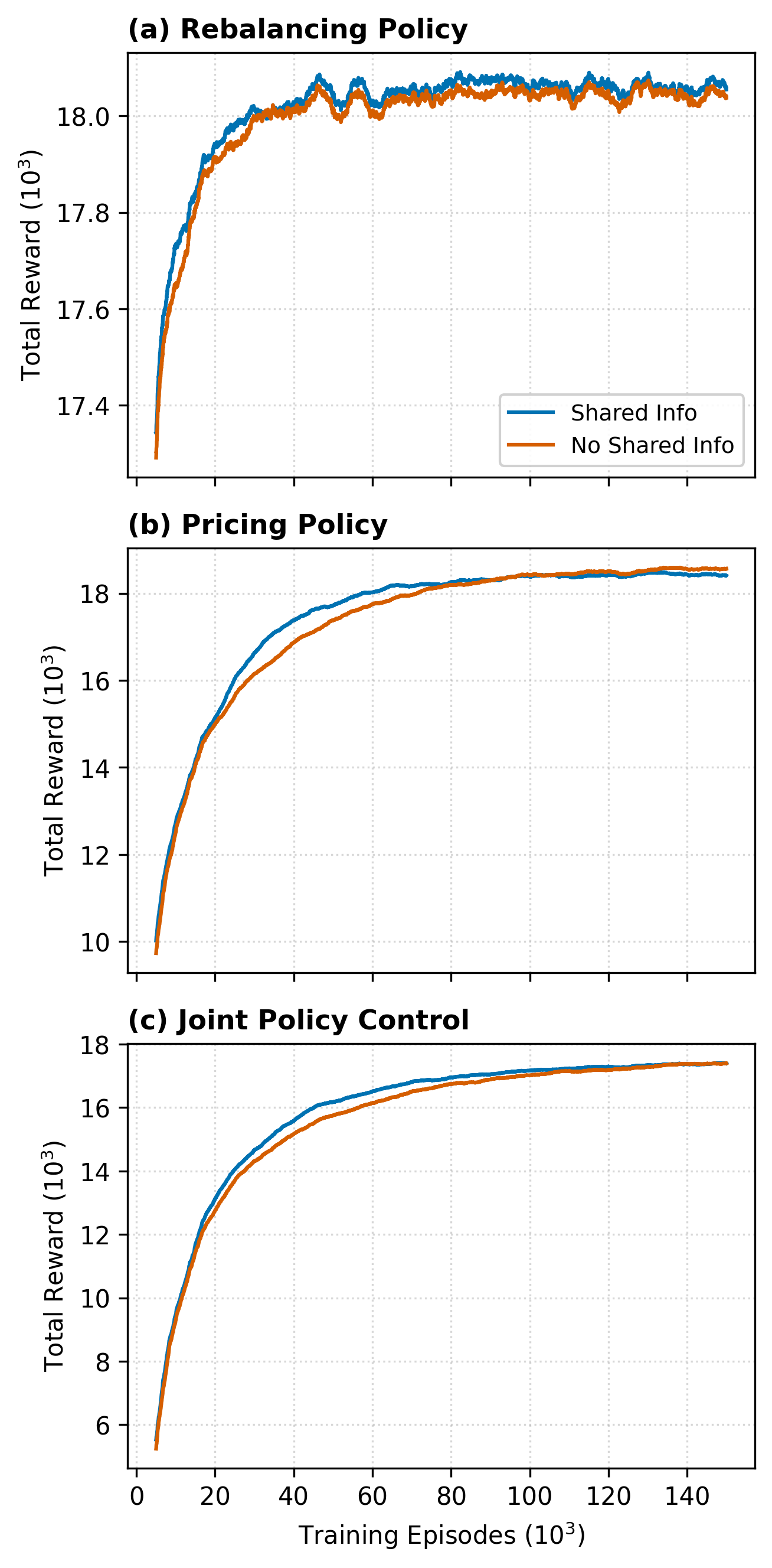}
\caption{Convergence dynamics comparing scenarios with and without competitor price visibility: (a) Rebalancing, (b) Pricing, (c) Joint. Curves are smoothed over 30 episodes; the first 5,000 episodes are excluded for clarity. Note that rewards are training rewards based on sampling from the policy distributions and may therefore be lower than test rewards.}
\label{fig:convergence_info_sharing}
\end{figure}
\vspace{2cm}

\clearpage

\section*{Regional Wage Heterogeneity}
\begin{table}[H]
\centering
\label{tab:wage_heterogeneity_results}
\begin{tabular}{lc}
\toprule
Metric & Joint Policy \\
\midrule
Total Reward & 23564.6 (341.1) \\
Reward Operator 0 & 11659.3 (206.4) \\
Reward Operator 1 & 11905.3 (248.4) \\
\cmidrule(lr){1-2}
Total Rebalancing Costs & 2661.3 (87.2) \\
Rebalancing Costs Operator 0 & 1359.8 (52.2) \\
Rebalancing Costs Operator 1 & 1301.5 (57.0) \\
\cmidrule(lr){1-2}
Total Rebalance Trips & 609.7 (21.0) \\
Rebalance Trips Operator 0 & 315.0 (10.9) \\
Rebalance Trips Operator 1 & 294.7 (15.0) \\
\cmidrule(lr){1-2}
Total Served Demand & 2990.9 (28.1) \\
Served Demand Operator 0 & 1480.6 (18.9) \\
Served Demand Operator 1 & 1510.3 (21.1) \\
\cmidrule(lr){1-2}
Price Operator 0 & 1.32 (0.00) \\
Price Operator 1 & 1.31 (0.00) \\
\cmidrule(lr){1-2}
Wait/mins Operator 0 & 1.42 (0.07) \\
Wait/mins Operator 1 & 1.46 (0.09) \\
\cmidrule(lr){1-2}
Queue Operator 0 & 6.40 (1.01) \\
Queue Operator 1 & 5.97 (1.34) \\
\cmidrule(lr){1-2}
Total Demand & 3567.0 (46.7) \\
Demand Operator 0 & 1762.8 (24.5) \\
Demand Operator 1 & 1804.2 (33.7) \\
\cmidrule(lr){1-2}
Average Wage & 25.8 (0.1) \\
\bottomrule
\end{tabular}
\end{table}

\begin{figure}[H]
\centering
\includegraphics[width=0.40\textwidth]{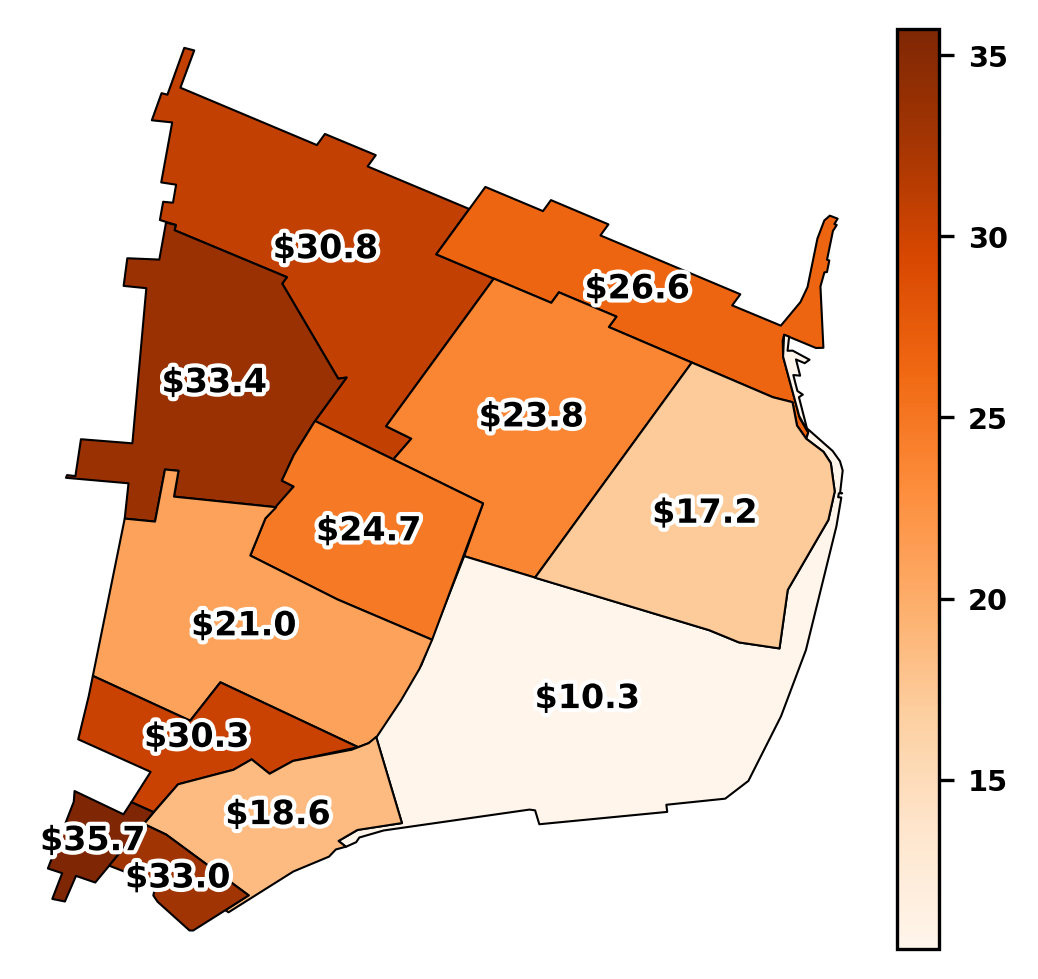}
\caption{Average hourly passenger wage distribution across regions in NYC Manhattan South under regional income heterogeneity.}
\label{fig:wage_distribution}
\end{figure}

\begin{figure}[H]
  \centering
  \subfloat[Operator 0]{
    \includegraphics[width=0.45\columnwidth]{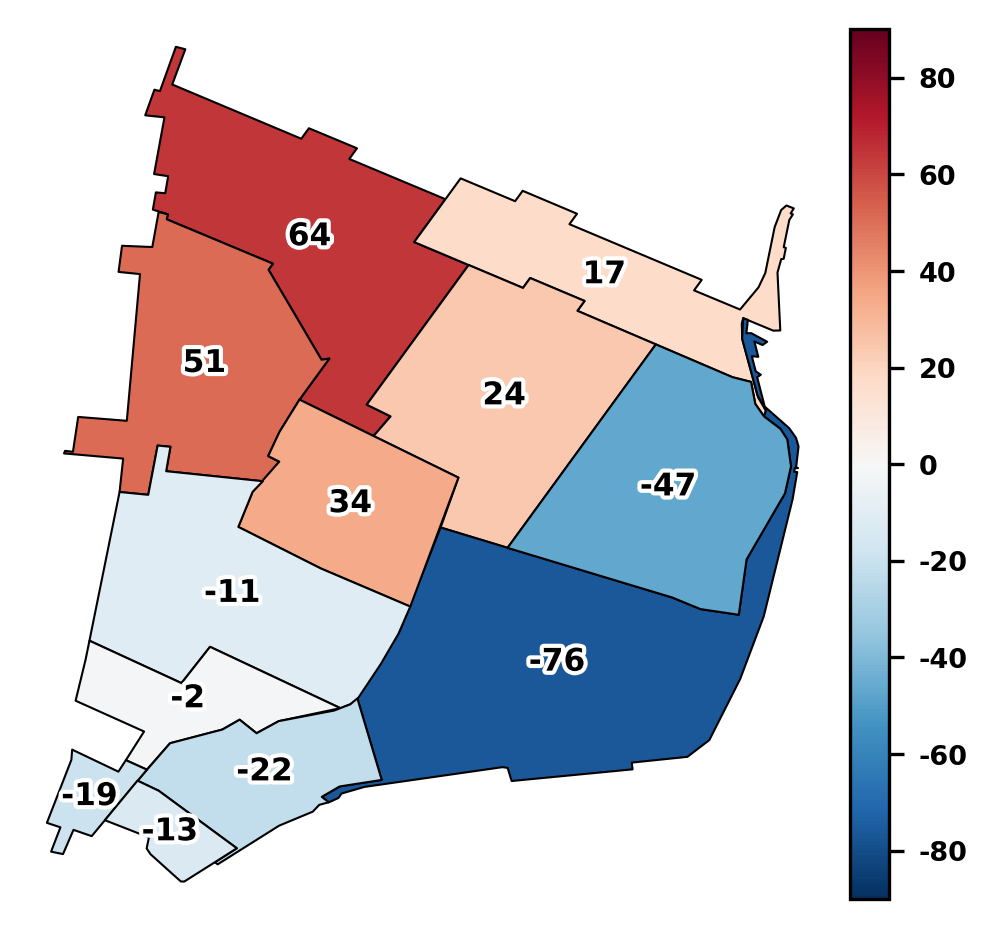}
    \label{fig:reb-dep-a0}
  }\hfill
  \subfloat[Operator 1]{
    \includegraphics[width=0.45\columnwidth]{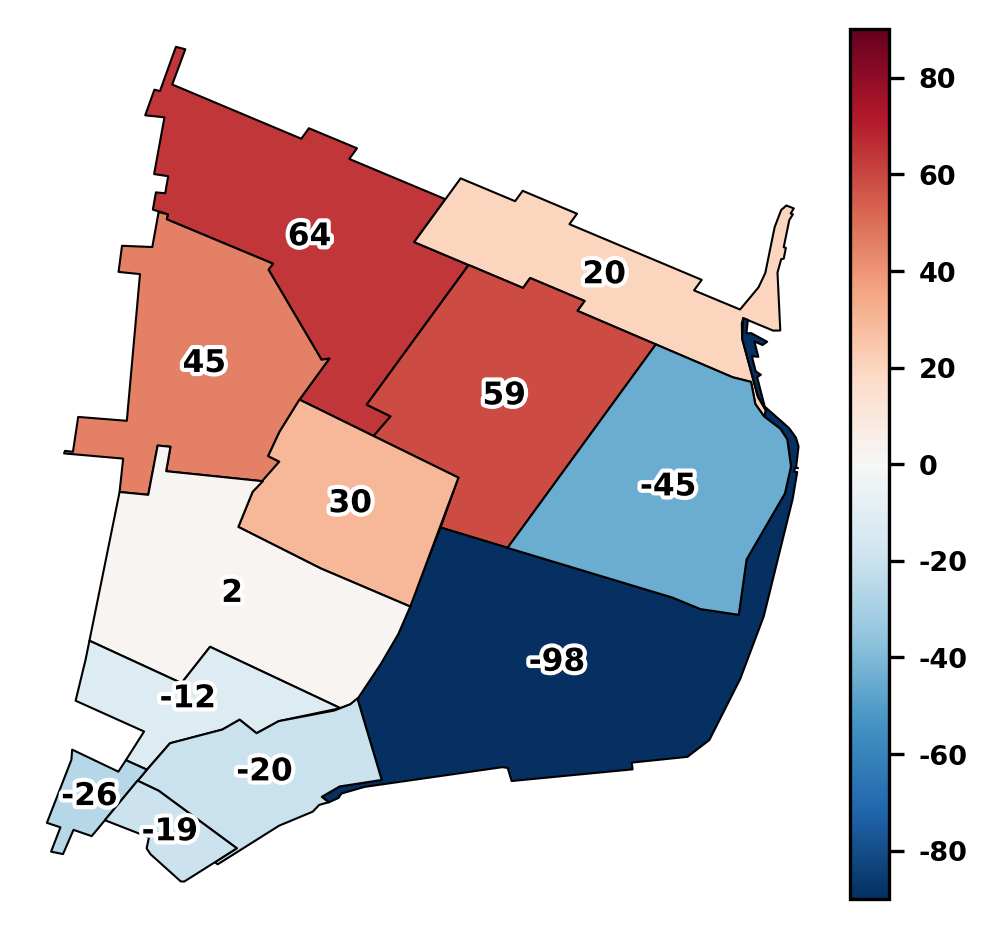}
    \label{fig:reb-dep-a1}
  }
  \caption{Net rebalancing flows under regional income heterogeneity, showing cumulative vehicle movements across all time steps. Red = net receiver, blue = net sender.}
  \label{fig:rebalancing_flows_wage}
\end{figure}

\begin{figure}[H]
  \centering
  \subfloat[Operator 0]{
    \includegraphics[width=0.45\columnwidth]{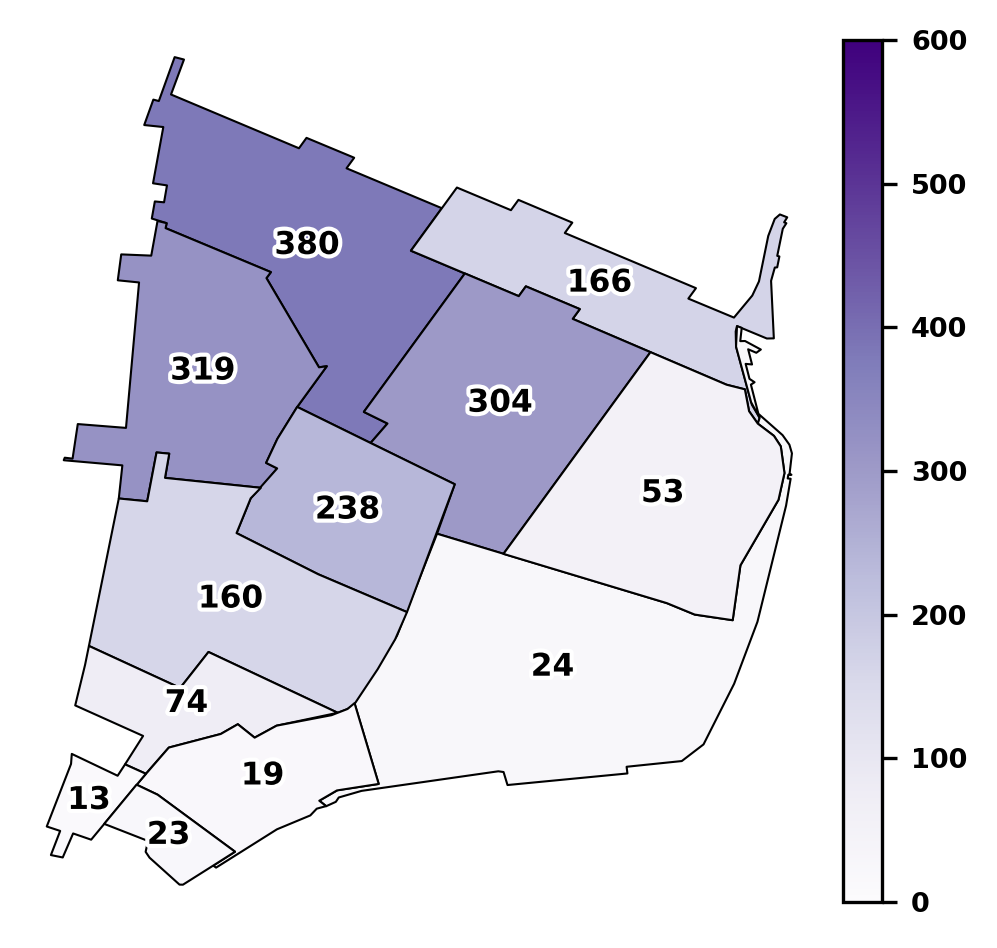}
    \label{fig:reb-dep-a0}
  }\hfill
  \subfloat[Operator 1]{
    \includegraphics[width=0.45\columnwidth]{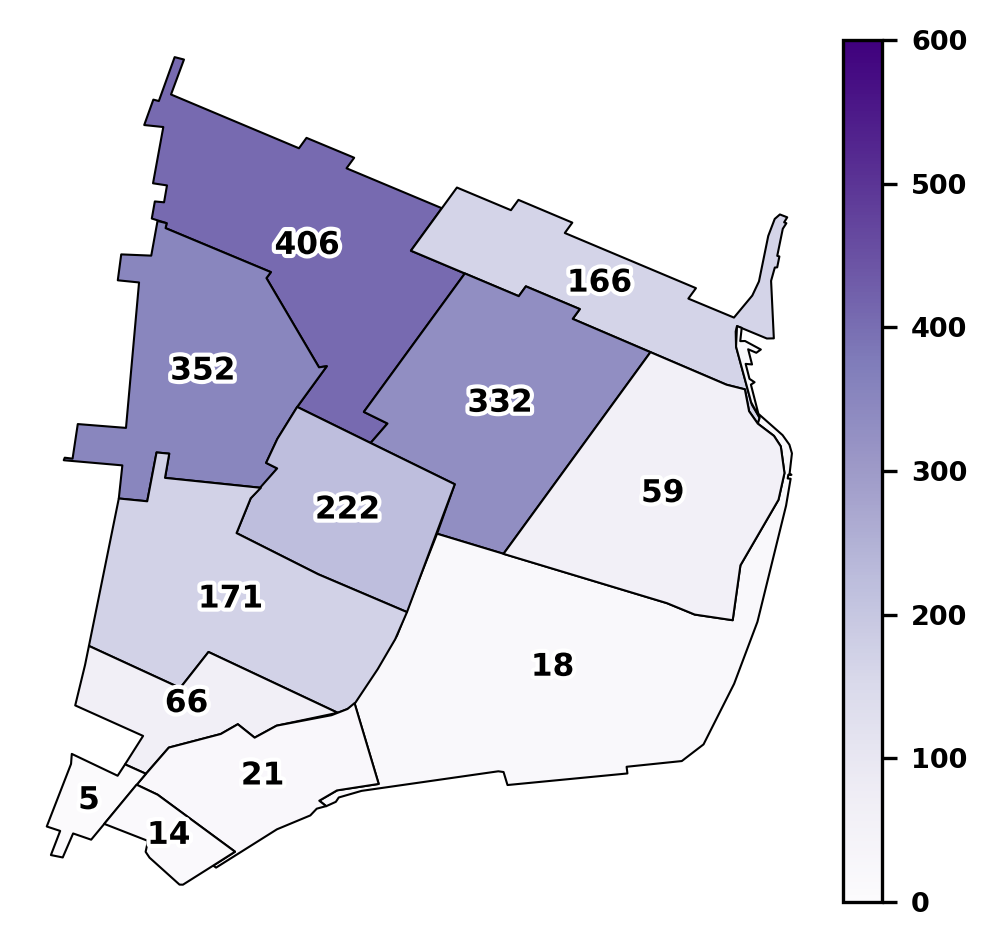}
    \label{fig:reb-dep-a1}
  }
  \caption{Total demand originating from each region under regional income heterogeneity.}
  \label{fig:demand_wage}
\end{figure}

\begin{figure}[H]
  \centering
  \subfloat[Operator 0]{
    \includegraphics[width=0.45\columnwidth]{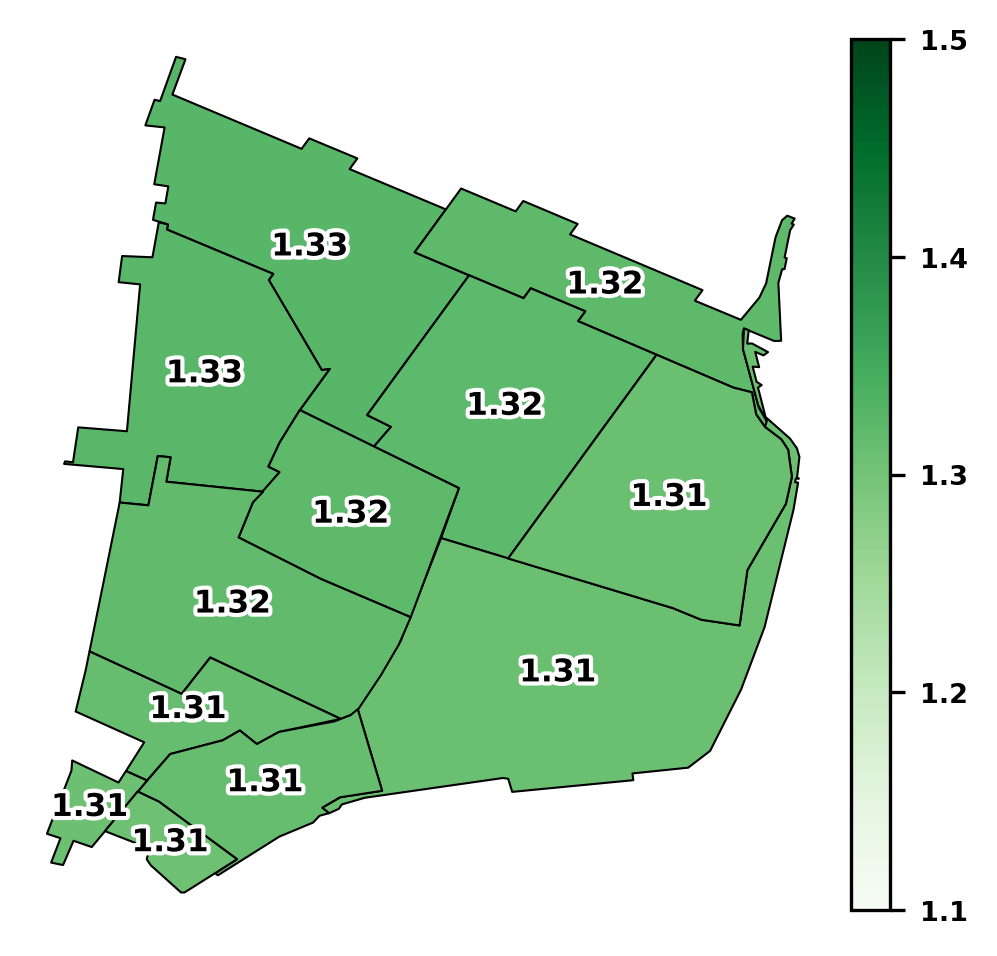}
    \label{fig:reb-dep-a0}
  }\hfill
  \subfloat[Operator 1]{
    \includegraphics[width=0.45\columnwidth]{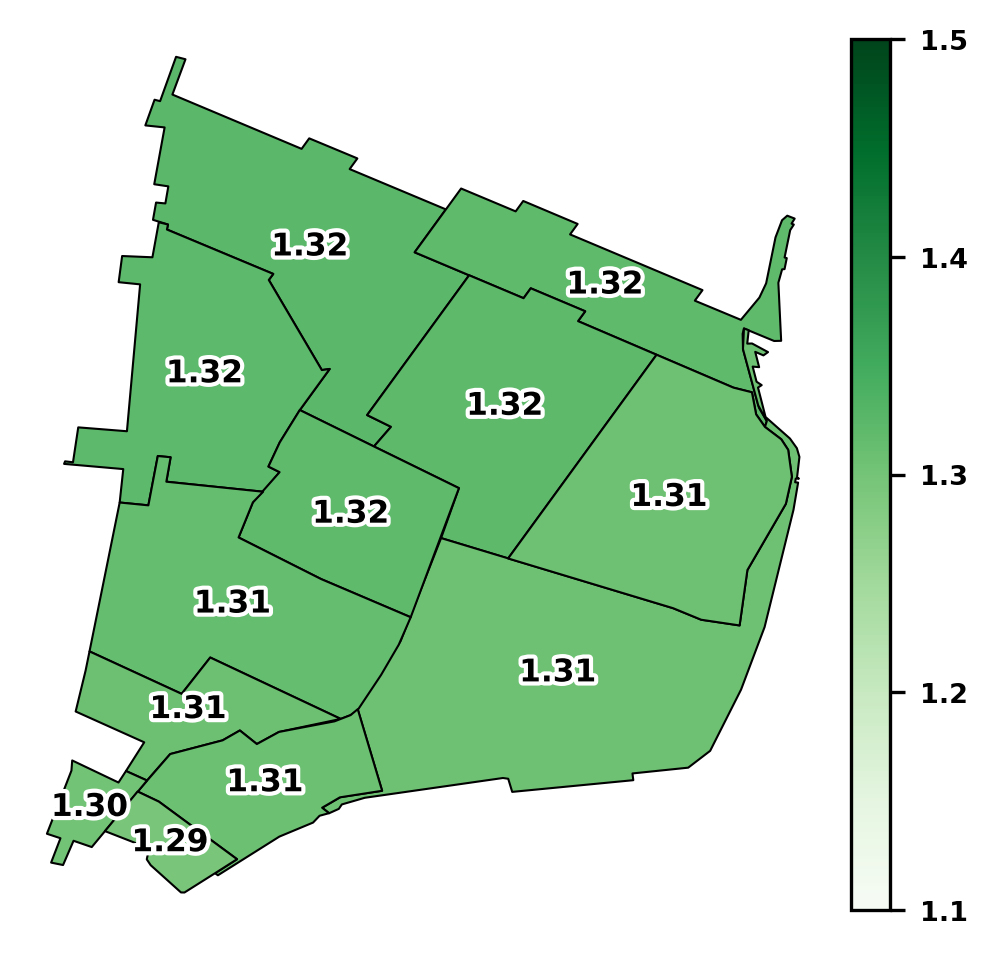}
    \label{fig:reb-dep-a1}
  }
  \caption{Average pricing scalars per region across all time steps under regional income heterogeneity.}
  \label{fig:pricing_wage}
\end{figure}
    
\end{document}